\def\arxivbuild{1}
\documentclass[letterpaper]{article}
\ifdefined\arxivbuild
  \usepackage[preprint]{aaai2027}
\else
  \usepackage[submission]{aaai2027}
\fi
\usepackage[hyphens]{url}
\usepackage{graphicx}
\usepackage{adjustbox}
\usepackage{natbib}
\usepackage{caption}
\usepackage{amsfonts}
\usepackage{amssymb}
\usepackage{amsmath}
\usepackage{mathtools}
\usepackage{algorithm}
\usepackage{algorithmic}
\usepackage{booktabs}
\usepackage{float}
\usepackage{nicefrac}
\usepackage{xcolor}
\usepackage{subcaption}

\newcommand{\norm}[1]{\left\lVert#1\right\rVert}

\title{Policy Gradient Steering: Interventions from Behavioral Objectives}
\ifdefined\arxivbuild
  \author{
    Yoann Poupart\textsuperscript{\rm 1},
    Aurélie Beynier\textsuperscript{\rm 1},
    Nicolas Maudet\textsuperscript{\rm 1}
  }
  \affiliations{
    \textsuperscript{\rm 1}LIP6, Sorbonne University\\
    Paris, France\\
    yoann.poupart@lip6.fr
  }
\else
  \author{Anonymous Submission}
  \affiliations{}
\fi

\begin{document}

\maketitle

\begin{abstract}
    Activation steering has emerged in large language models as a lightweight alternative for dynamically changing a model’s behavior at inference time. However, we show that existing steering methods fail to steer even a simple policy in a two-route gridworld environment. To address this limitation, we propose Policy Gradient Steering (PGS), which formulates steering as a reinforcement learning problem. PGS accumulates gradients of a temporary behavioral objective over a small set of rollouts or demonstrations to construct a removable task vector. We first demonstrate the calibration and reversibility of PGS in a two-route gridworld environment. Using chess puzzles, we then evaluate independently fitted PGS vectors both in isolation and in combination, finding that compatible tactical objectives accumulate constructively. Finally, in competitive football, we show that PGS can alter specific team behaviors and that its effects transfer across opponents. Together, these results show that policy gradients provide a natural interface for constructing temporary and composable behavioral adaptations across diverse decision-making domains.
\end{abstract}

\section{Introduction}
\label{rethink:sec:introduction}

Trained policies are typically optimized for objectives fixed during training.
Although some policies expose controls conditioned on preferences or returns
\cite{Abels2019DynamicWeights,Chen2021DecisionTransformer}, a fixed deployed
policy may provide no such interface. After deployment, however, we may want to
impose a new behavioral preference, such as favoring one valid strategy over
another, without retraining or permanently modifying the base policy. The
challenge is to turn a small set of trajectories scored under this temporary
objective into a controllable intervention that can be applied, scaled, and
removed at inference time.

Existing activation-steering methods commonly construct interventions from
representations associated with contrasting examples or rollout outcomes
\cite{rimsky-etal-2024-steering,miao2026coast}. This construction can fail for feed-forward policies such as MLPs, whose pre-decision activations depend only on the current observation. When trajectories diverge from the same observation, their activations differ only after they reach different states. As our two-route gridworld shows, the resulting contrast may encode downstream state visitation without controlling the branching action.

To address this limitation, we introduce Policy Gradient Steering (PGS), which
uses policy gradients to translate behavioral feedback on trajectories into an
additive activation intervention. Rather than contrasting representations
reached after outcomes diverge, PGS assigns credit to the actions that produced
those outcomes and aggregates the resulting activation gradients into a single
steering vector. The vector can be scaled, composed, and removed at inference
time while the base policy remains frozen.

\begin{figure}[t]
  \raggedleft
  \includegraphics[width=0.95\linewidth]{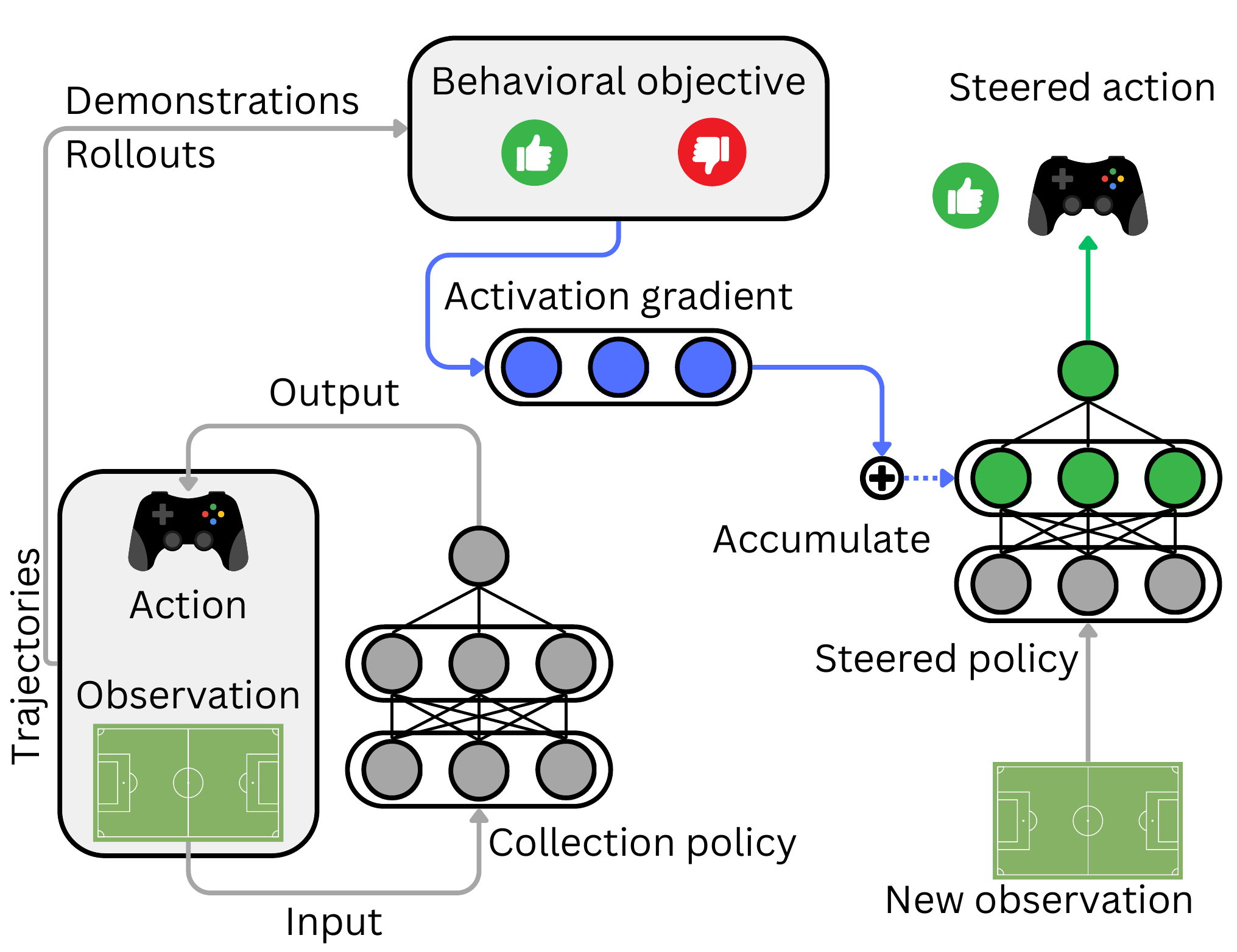}
  \caption{Overview of the policy gradient steering method. Activation gradients are accumulated from a small batch of scored trajectories to construct an additive steering vector.}
  \label{rethink:fig:overview}
\end{figure}

PGS occupies a middle ground between activation steering and policy
fine-tuning. Like contrastive activation methods, it provides lightweight
inference-time control without modifying the base policy
\cite{rimsky-etal-2024-steering,miao2026coast}. Like policy fine-tuning and
learned adaptations such as ReFT and LoRA, however, it derives the intervention
by optimizing the behavioral objective
\cite{Xie2021PolicyFinetuning,wu2024reft,hu2021loralowrankadaptationlarge}. This
raises a further question: whether independently constructed behavioral
interventions can be combined without retraining, as explored for parameter and
activation interventions in prior work
\cite{Ilharco2022EditingMW,Pfeiffer2021AdapterFusion,
Stolfo2024InstructionSteering}. We study this compositional property directly
in chess.

We evaluate PGS across three settings that expose distinct challenges. A
two-route gridworld isolates the action-credit failure of outcome-based
contrasts. Chess tests whether independently fitted tactical interventions
retain their effects under composition and across policies of different skill
levels. Competitive football extends the evaluation to multi-agent policies,
where an intervention's effect may depend on the controlled policy and its
strategic context.

\paragraph{Contributions.}
\begin{enumerate}
  \item Policy Gradient Steering, a method that converts arbitrary scalar
  feedback on trajectories into a removable activation intervention by
  assigning credit to the policy's actions rather than to post-outcome
  representations.
  \item A controlled gridworld analysis showing that outcome-based activation
  contrasts can encode states reached after a decision without controlling the
  decision itself, supported by comparisons with contrastive steering methods
  and learned adaptations.
  \item An empirical evaluation across chess and competitive football examining
  how compact behavioral interventions compose across objectives and how their
  effects depend on the policy context.
\end{enumerate}

\section{Background}
\label{rethink:sec:background}

\subsection{Post-Training Behavioral Adaptation}

For a trained policy $\pi_{\theta_0}(a\mid o)$, we first consider behavioral
adaptations that rely on additional optimization.

\paragraph{Fine-tuning and parameter task vectors.}
In reinforcement learning, policy fine-tuning has been studied as online learning with access to a reference
policy~\cite{Xie2021PolicyFinetuning}. Our fine-tuning baseline updates policy weights by continuing optimization from the pretrained parameters under the newly specified behavioral objective.
When the base and adapted policies share an architecture,
task arithmetic represents the effect of this optimization as a parameter
difference
\cite{Ilharco2022EditingMW}:
\begin{equation}
  \tau_b=\theta_b-\theta_0,
\end{equation}
where $\theta_b$ denotes the fine-tuned weights. The resulting task vector can
be applied with strength $\alpha$ as
\begin{equation}
  \theta(\alpha)=\theta_0+\alpha\tau_b.
\end{equation}
This representation permits the adaptation to be scaled, negated, or combined
with other parameter-space adaptations.

\paragraph{Low-rank tuning.}
Low-rank tuning restricts adaptation to a compact set of task-specific
parameters. LoRA learns low-rank updates to the policy weights
\cite{hu2021loralowrankadaptationlarge}, whereas ReFT freezes the weights and
learns a low-rank intervention on hidden representations~\cite{wu2024reft}.
In our LoRA baseline, a task-specific adapter is fitted and its effective
weight-space update is treated as a scalable, composable task-specific
adaptation, following the task-arithmetic treatment of parameter
deltas~\cite{Ilharco2022EditingMW}.

\paragraph{Preference adaptation.}
One way to support changing preferences at runtime is to train a policy that
conditions its behavior on an explicit preference variable
\cite{Abels2019DynamicWeights}. Successor features similarly separate expected
reward features from task-specific weights, allowing adaptation to a new linear
reward when the feature basis is known
\cite{Dayan1993Successor,Barreto2018SuccessorFeatures,Borsa2019USFA}. When a
fixed policy exposes no such interface, an alternative is to construct an
adaptation after training. In language models, DPO adapts a policy from
preferred--rejected pairs~\cite{Rafailov2023DPO}, and Preference Vectors derive a scalable
parameter-space adaptation direction from such feedback
\cite{Liang2026PreferenceVectors}.

\subsection{Activation Steering Baselines}
\label{rethink:sec:background-contrastive-baselines}

These methods construct activation interventions from positive and negative
examples without using behavioral-return gradients through the policy.

\paragraph{Contrastive activation addition (CAA).}
Given positive and negative example sets $\mathcal D_+$ and $\mathcal D_-$,
CAA adds their mean activation difference at layer $l$
\cite{rimsky-etal-2024-steering}:
\begin{equation}
  h^l \leftarrow h^l+\alpha\left(
  \mathbb E_{\mathcal D_+}[h^l]
  -\mathbb E_{\mathcal D_-}[h^l]
  \right).
  \label{rethink:eq:background-caa}
\end{equation}
where $h^l$ denotes the activation at layer $l$, and $\alpha$ controls the
intervention strength.

\paragraph{K-Steer.}
K-Steer trains a nonlinear multi-label classifier to recognize behavioral
attributes from hidden activations~\cite{oozeer2025ksteering}. Given a
classifier $c_\phi$ and a loss $\mathcal L$ that rewards desired attributes and
penalizes attributes to avoid, it updates the current activation as
\begin{equation}
  h^l \leftarrow h^l
  -\alpha\nabla_{h^l}\mathcal L\!\left(c_\phi(h^l)\right).
  \label{rethink:eq:background-k-steer}
\end{equation}

\paragraph{COAST.}
COAST, introduced for vision-language-action models, fits conceptors
$C_{\mathrm{success}}$ and $C_{\mathrm{failure}}$ from activations collected
during successful and failed rollouts~\cite{miao2026coast}. It applies the
resulting contrastive conceptor as
\begin{equation}
  h^l\leftarrow
  \left[(1-\alpha)I
  +\alpha\left(C_{\mathrm{success}}
  \wedge(I-C_{\mathrm{failure}})\right)\right]h^l,
  \label{rethink:eq:background-coast}
\end{equation}
where $\wedge$ denotes conceptor intersection.

\section{Policy Gradient Steering}
\label{rethink:sec:rl-optimization}

\subsection{Problem Formulation}

Let $\pi(a\mid o)$ be a pretrained policy that maps an observation $o$ to a
distribution over actions. After training, we introduce a temporary behavioral
objective through a scalar reward $r_b$.

For a trajectory $\tau=(o_0,a_0,\ldots,o_T,a_T)$, the behavioral return
associated with the decision at time $t$ is
\begin{equation}
  G_t=\sum_{k=t}^{T}\gamma^{k-t}r_b(o_k,a_k),
  \label{rethink:eq:behavioral-return}
\end{equation}
where $\gamma\in[0,1]$ controls how later outcomes are assigned to earlier
decisions.

Given a small dataset of reward-scored trajectories
$\mathcal D=\{\tau_1,\ldots,\tau_N\}$ collected by a behavior policy $\mu$,
the steering problem is to obtain a controlled policy $\widetilde{\pi}$ that
maximizes
\begin{equation}
  \max_{\widetilde{\pi}}\;
  \mathbb E_{\tau\sim\widetilde{\pi}}\!\left[G(\tau)\right].
  \label{rethink:eq:steering-objective}
\end{equation}
We write $G(\tau)=G_0$ for the trajectory return.
Classical contrastive steering formulation is recovered by splitting
$\mathcal D$ equally into positive and negative trajectories and assigning
each trajectory the scalar reward $r_b(\tau)=+1$ or $r_b(\tau)=-1$,
respectively.

\subsection{Constructing the PGS Vector}

At a chosen activation $h$, PGS applies the likelihood-ratio construction
underlying REINFORCE and the policy-gradient theorem
\cite{Williams1992,Sutton1999PolicyGradient}: it accumulates gradients of the
observed actions weighted by their behavioral returns,
\begin{equation}
  v_{\mathrm{PGS}}
  \coloneqq
  \mathbb E_{\mathcal D}
  \left[
    \rho_t\bigl(G_t-b(o_t)\bigr)
    \nabla_h\log\pi(a_t\mid o_t)
  \right],
  \label{rethink:eq:pgs-vector}
\end{equation}
Unlike policy-gradient fine-tuning, Equation~\ref{rethink:eq:pgs-vector} differentiates with respect to the selected activation and averages these decision-level gradients into a single state-independent offset, which can be stored and applied without modifying the policy weights.
Once its steering strength $\alpha$ has been calibrated, inference applies the
resulting vector as
\begin{equation}
  h\leftarrow h+\alpha v_{\mathrm{PGS}}.
  \label{rethink:eq:pgs-application}
\end{equation}
For data collected by behavior policy $\mu$, the per-decision importance
ratio is~\cite{Degris2012}
\begin{equation}
  \rho_t=
  \frac{\pi(a_t\mid o_t)}{\mu(a_t\mid o_t)}.
  \label{rethink:eq:importance-ratio}
\end{equation}
It is treated as fixed when computing the gradient and equals one for on-policy
data. For off-policy data, this simple per-decision ratio
corrects the action distribution only at the observations sampled under
$\mu$; it does not in general correct the mismatch between the state
distributions of the behavior and target policies. We therefore treat it as
an approximate off-policy construction rather than a generally unbiased
estimator of Equation~\ref{rethink:eq:steering-objective}. We use an
action-independent baseline $b(o_t)$ to reduce return variance without
changing the expected on-policy gradient~\cite{Williams1992}.

\subsection{Steering Algorithm}

Algorithm~\ref{rethink:alg:pgs} summarizes fitting. Inference applies the
fitted intervention in the policy's forward pass using
Equation~\ref{rethink:eq:pgs-application}. To calibrate its strength, we use the
activation-space Fisher matrix
\begin{equation}
  F=
  \mathbb E_{\substack{o\sim\mathcal D\\a\sim\pi(\cdot\mid o)}}
  \left[
    \nabla_h\log\pi(a\mid o)
    \nabla_h\log\pi(a\mid o)^\top
  \right].
  \label{rethink:eq:activation-fisher}
\end{equation}
Under a local quadratic approximation to the action KL
(Appendix~\ref{rethink:app:activation-sensitivity}), the coefficient for a
policy-change budget $\varepsilon$ is
\begin{equation}
  \alpha=
  \sqrt{\frac{2\varepsilon}
  {v_{\mathrm{PGS}}^\top Fv_{\mathrm{PGS}}}}.
  \label{rethink:eq:pgs-scale}
\end{equation}
We calibrate the intervention from the scalar directional curvature
$v_{\mathrm{PGS}}^\top Fv_{\mathrm{PGS}}$.
We discuss trust-region calibration and natural-gradient preconditioning as
future extensions in Section~\ref{rethink:sec:discussion}.

\begin{algorithm}[ht]
\caption{Policy Gradient Steering}
\label{rethink:alg:pgs}
\textbf{Input}: policy $\pi$, trajectories $\mathcal D$, behavioral reward
$r_b$, activation site $l$, KL budget $\varepsilon$\\
\textbf{Output}: PGS intervention $(v_{\mathrm{PGS}},\alpha)$
\begin{algorithmic}[1]
  \STATE Compute returns $G_t$ from $r_b$
  \STATE Estimate the baseline $b(o_t)$
  \STATE Accumulate $v_{\mathrm{PGS}}$ at site $l$ using
    Equation~\ref{rethink:eq:pgs-vector}
  \STATE Estimate the directional curvature
    $v_{\mathrm{PGS}}^\top Fv_{\mathrm{PGS}}$
  \STATE $\alpha\leftarrow
    \sqrt{2\varepsilon/
    (v_{\mathrm{PGS}}^\top Fv_{\mathrm{PGS}})}$
  \RETURN $(v_{\mathrm{PGS}},\alpha)$
\end{algorithmic}
\end{algorithm}

\section{Limits of Existing Activation Steering}
\label{rethink:sec:activation-limits}

\subsection{Gridworld Protocol}

\begin{figure}[H]
  \centering
  \begin{subfigure}[b]{0.47\linewidth}
    \centering
    \includegraphics[width=\linewidth]{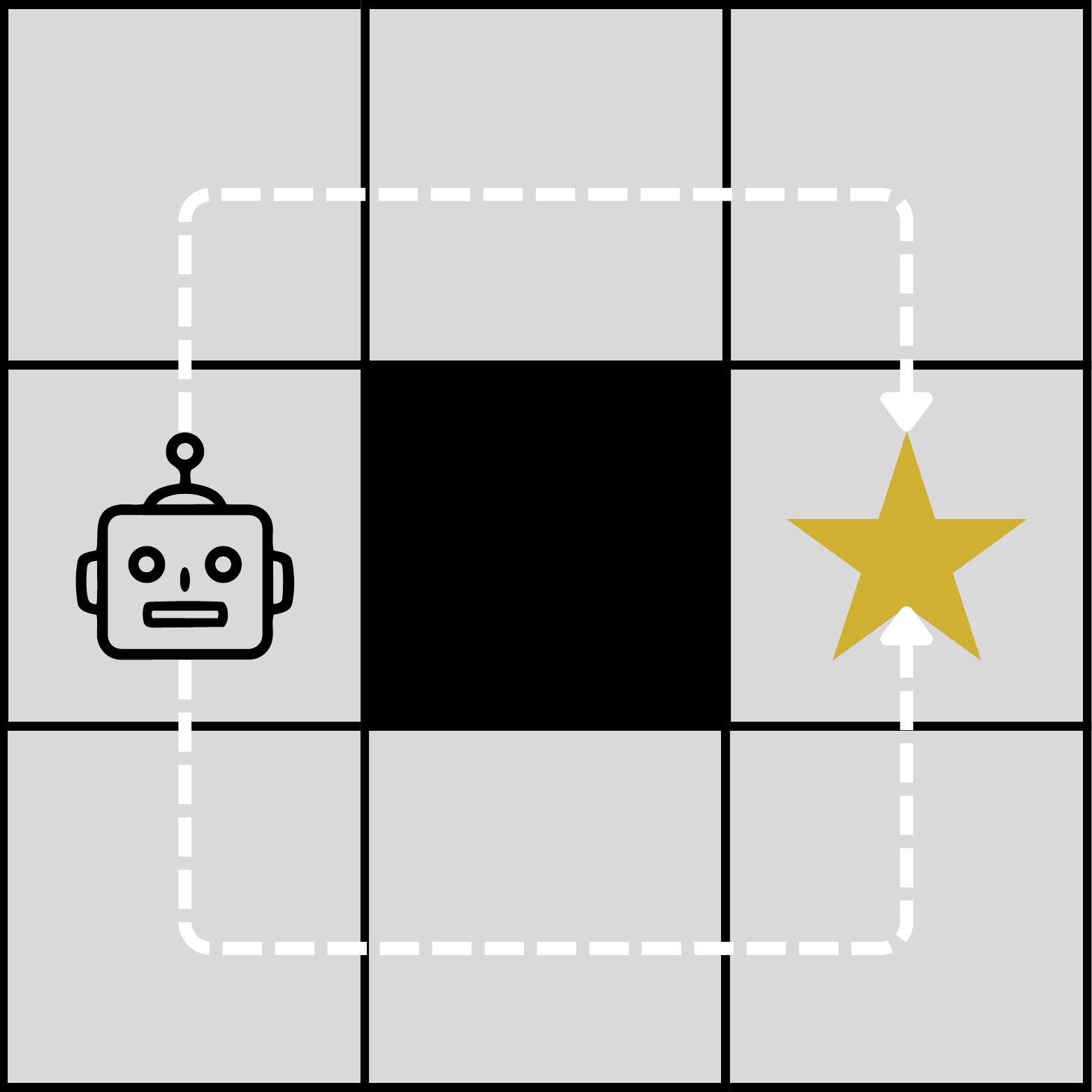}
    \caption{Navigation training}
    \label{rethink:fig:gridworld-navigation-task}
  \end{subfigure}\hfill
  \begin{subfigure}[b]{0.47\linewidth}
    \centering
    \includegraphics[width=\linewidth]{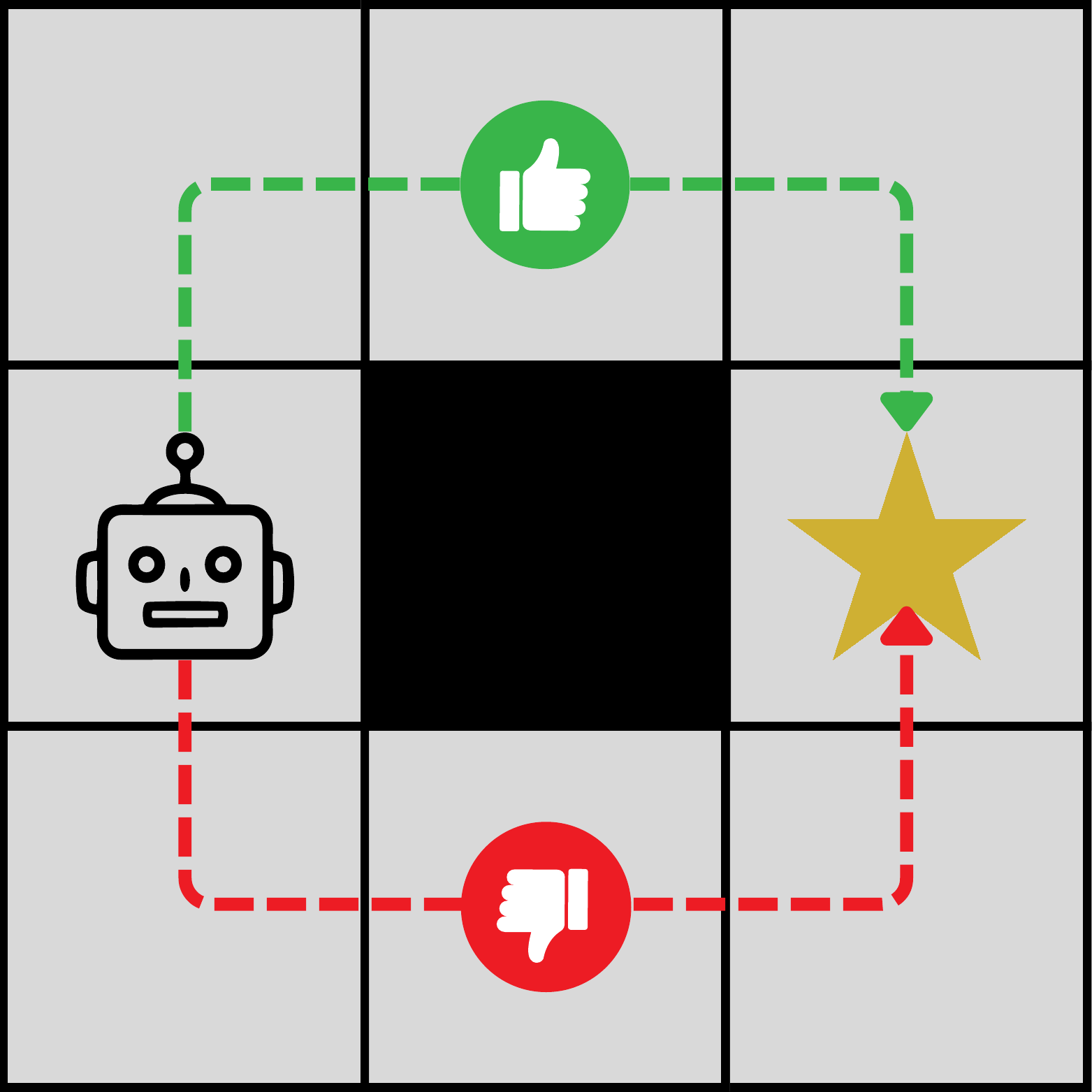}
    \caption{Route preference}
    \label{rethink:fig:gridworld-route-preference-task}
  \end{subfigure}
  \caption{Two-route Gridworld used to isolate behavioral preference from task
  completion. In (a), the policy learns to navigate from the start square to the
  goal. In (b), the upper route is preferred and the lower route is
  discouraged.}
  \label{rethink:fig:gridworld-environment}
\end{figure}

Figure~\ref{rethink:fig:gridworld-environment} shows the two-route environment.
The policy is trained only to navigate from the start state to the goal; route
preference is introduced after training.

For the trained MLP policy, layer $l$ has pre-activation and post-ReLU
representation
\begin{align}
  h_{\mathrm{pre}}^l
    &=W^l h_{\mathrm{post}}^{l-1}+b^l,
    \label{rethink:eq:activation-preactivation}\\
  h_{\mathrm{post}}^l
    &=\operatorname{ReLU}(h_{\mathrm{pre}}^l),
    \label{rethink:eq:activation-relu}
\end{align}
with $h_{\mathrm{post}}^0=o$. An activation intervention inserts a function
$g_l$ before the remaining policy layers while leaving $W^l$ and $b^l$ fixed.
CAA, COAST, K-Steer, and the random activation baseline intervene at
$h_{\mathrm{post}}^2$. PGS jointly fits offsets at
$h_{\mathrm{post}}^1$ and $h_{\mathrm{post}}^2$ and normalizes their
concatenation as one direction. These sites are fixed before fitting, while
intervention strength is selected using validation data only. PGS uses one
gradient update and fine-tuning uses three updates on the same fixed batch.

To adapt the contrastive baselines to route steering, we treat activations from
successful target-route trajectories as positive and those from successful
alternative-route trajectories as negative, excluding failures. Every valid
timestep inherits its trajectory label, and we do not balance the sets by
subsampling. CAA contrasts the two set means, COAST fits a conceptor to each
set, and K-Steer uses a class-balanced classifier loss.

\subsection{Gridworld Results}

\paragraph{Activation steering.}
We first fit each method using trajectories sampled from the frozen policy.
Existing activation steering baselines remain close to the unsteered route
distribution. PGS instead shifts the route preference in the requested
direction while preserving path efficiency.

\begin{table}[H]
  \centering
  \small
  \setlength{\tabcolsep}{4.5pt}
  \adjustbox{max width=\columnwidth}{%
  \begin{tabular}{lcccc}
    \toprule
    Method & Selected $\alpha$ & Top prob. $(+)$ $\uparrow$ & Bottom prob. $(-)$ $\uparrow$ & Path length $(+)$ $\downarrow$ \\
    \midrule
    No intervention & -- & $0.43 \pm 0.06$ & $0.57 \pm 0.06$ & $4.36 \pm 0.07$ \\
    Random activation & $0.23 \pm 0.30$ & $0.45 \pm 0.06$ & $0.57 \pm 0.08$ & $4.36 \pm 0.07$ \\
    CAA & $0.11 \pm 0.08$ & $0.41 \pm 0.07$ & $0.52 \pm 0.08$ & $\mathbf{4.33 \pm 0.08}$ \\
    COAST & $0.23 \pm 0.19$ & $0.49 \pm 0.02$ & $0.51 \pm 0.02$ & $4.61 \pm 0.25$ \\
    K-Steer & $0.09 \pm 0.09$ & $0.45 \pm 0.06$ & $0.56 \pm 0.06$ & $\underline{4.34 \pm 0.07}$ \\
    Fine-tuning & $0.20 \pm 0.09$ & $\mathbf{0.86 \pm 0.13}$ & $\mathbf{0.92 \pm 0.05}$ & $4.59 \pm 0.22$ \\
    \midrule
    PGS (ours) & $0.42 \pm 0.11$ & $\underline{0.82 \pm 0.07}$ & $\underline{0.83 \pm 0.07}$ & $4.54 \pm 0.13$ \\
    \bottomrule
  \end{tabular}%
  }
  \caption{Gridworld steering fitted from frozen-policy rollouts.}
  \label{tab:gridworld-rollouts}
\end{table}

\paragraph{Parameterization.}
We next compare intervention parameterizations without selecting application
strength from behavioral returns. For each fitted direction, we estimate its
local Fisher curvature on the fit states and choose the scale for a shared
target action-KL budget. PGS is the strongest compact realization, while
fine-tuning is strongest overall. ReFT and LoRA produce smaller route shifts,
and all parameterizations retain comparable path lengths.
PGS, ReFT, and fine-tuning achieve the target KL closely. LoRA is less stable
because one fitted direction has near-zero local curvature. Across the tested
budgets, PGS's measured held-out KL closely tracks the requested budget
(separate appendix).

\begin{table}[H]
  \centering
  \small
  \setlength{\tabcolsep}{4.5pt}
  \adjustbox{max width=\columnwidth}{%
  \begin{tabular}{lrrcc}
    \toprule
    Method & Updates & Scalars $\downarrow$ & Top route $\uparrow$ & KL $\downarrow$ \\
    \midrule
    ReFT & 3 & $\underline{66}$ & $0.77 \pm 0.07$ & $\underline{0.097 \pm 0.006}$ \\
    LoRA & 3 & $76$ & $0.75 \pm 0.07$ & $0.138 \pm 0.097$ \\
    Fine-tuning & 3 & $484$ & $\mathbf{0.83 \pm 0.03}$ & $\mathbf{0.091 \pm 0.002}$ \\
    \midrule
    PGS (ours) & 1 & $\mathbf{32}$ & $\underline{0.80 \pm 0.04}$ & $0.097 \pm 0.004$ \\
    \bottomrule
  \end{tabular}%
  }
  \caption{Gridworld parameterization comparison at a common target KL budget of $0.1$. Application scales are computed from each fitted direction's local Fisher curvature without a behavioral scale sweep. Cells report mean $\pm$ sample standard deviation over fit seeds.}
  \label{tab:gridworld-parameterizations}
\end{table}

\subsection{Why Contrastive Steering Fails}

We isolate the failure using only two demonstrations: one successful
trajectory through each route. Even in this extreme low-data regime, PGS
shifts the route preference in both directions, whereas the contrastive
baselines remain close to the unsteered policy. Under this two-demonstration
fit, PGS produces the strongest top-route shift while fine-tuning is stronger
in the reverse direction.

\begin{table}[H]
  \centering
  \small
  \setlength{\tabcolsep}{4.5pt}
  \adjustbox{max width=\columnwidth}{%
  \begin{tabular}{lcccc}
    \toprule
    Method & Selected $\alpha$ & Top prob. $(+)$ $\uparrow$ & Bottom prob. $(-)$ $\uparrow$ & Path length $(+)$ $\downarrow$ \\
    \midrule
    No intervention & -- & $0.43 \pm 0.06$ & $0.57 \pm 0.06$ & $4.36 \pm 0.07$ \\
    Random activation & $0.23 \pm 0.30$ & $0.45 \pm 0.06$ & $0.57 \pm 0.08$ & $4.36 \pm 0.07$ \\
    CAA & $0.07 \pm 0.03$ & $0.40 \pm 0.07$ & $0.55 \pm 0.05$ & $\mathbf{4.33 \pm 0.08}$ \\
    COAST & $0.23 \pm 0.19$ & $0.49 \pm 0.02$ & $0.51 \pm 0.02$ & $4.61 \pm 0.25$ \\
    K-Steer & $0.24 \pm 0.42$ & $0.45 \pm 0.06$ & $0.59 \pm 0.08$ & $\underline{4.35 \pm 0.07}$ \\
    Fine-tuning & $0.15 \pm 0.09$ & $\underline{0.78 \pm 0.15}$ & $\mathbf{0.86 \pm 0.11}$ & $4.53 \pm 0.19$ \\
    \midrule
    PGS (ours) & $0.42 \pm 0.11$ & $\mathbf{0.82 \pm 0.07}$ & $\underline{0.83 \pm 0.08}$ & $4.51 \pm 0.17$ \\
    \bottomrule
  \end{tabular}%
  }
  \caption{Gridworld steering fitted from one demonstration per route.}
  \label{tab:gridworld-demonstrations}
\end{table}

The failure follows from the information available at the route decision. At
the branching observation, both demonstrations induce the same hidden
representation $h$. The decision-local contrast is therefore
\begin{equation}
  v_{\mathrm{CAA}}^{\mathrm{branch}}
    =h^{+}-h^{-}=h-h=0.
  \label{rethink:eq:gridworld-caa-zero}
\end{equation}
The fitted CAA baseline pools activations across all valid timesteps, so it may
recover a nonzero contrast from route-specific observations reached after the
decision. That contrast describes the consequences of the route choice rather
than the action that causes it. PGS can distinguish the two actions because
$\nabla_h\log\pi(a^{+}\mid o)$ and
$\nabla_h\log\pi(a^{-}\mid o)$ differ even when their pre-action
representations are identical.

\section{Evaluating Task Composition}
\label{rethink:sec:chess-composition}

\subsection{Experimental Setup and Metrics}

\begin{figure}[H]
  \centering
  \begin{subfigure}[b]{0.48\linewidth}
    \centering
    \includegraphics[width=\linewidth]{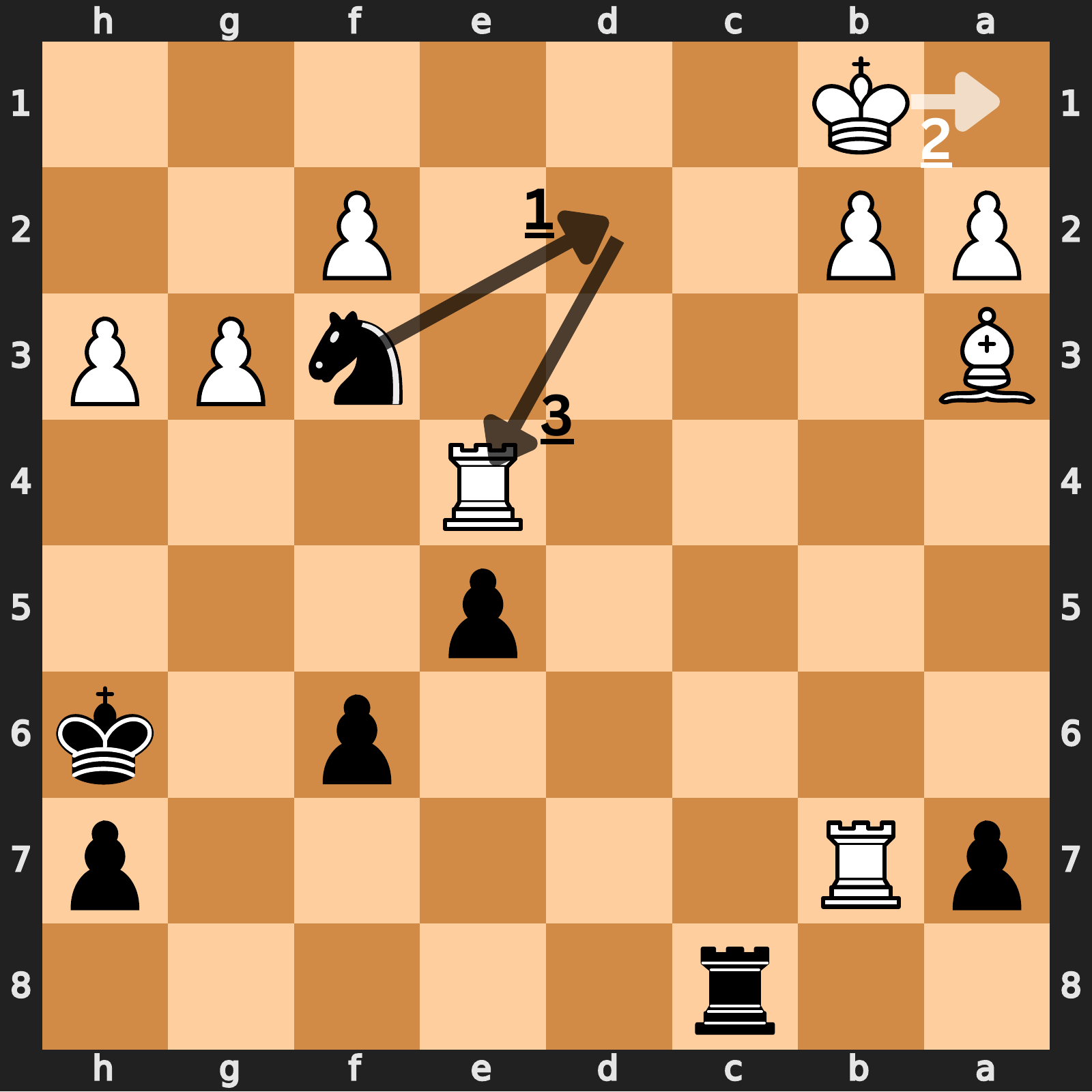}
    \caption{Fork puzzle}
    \label{rethink:fig:chess-fork-puzzle}
  \end{subfigure}\hfill
  \begin{subfigure}[b]{0.48\linewidth}
    \centering
    \includegraphics[width=\linewidth]{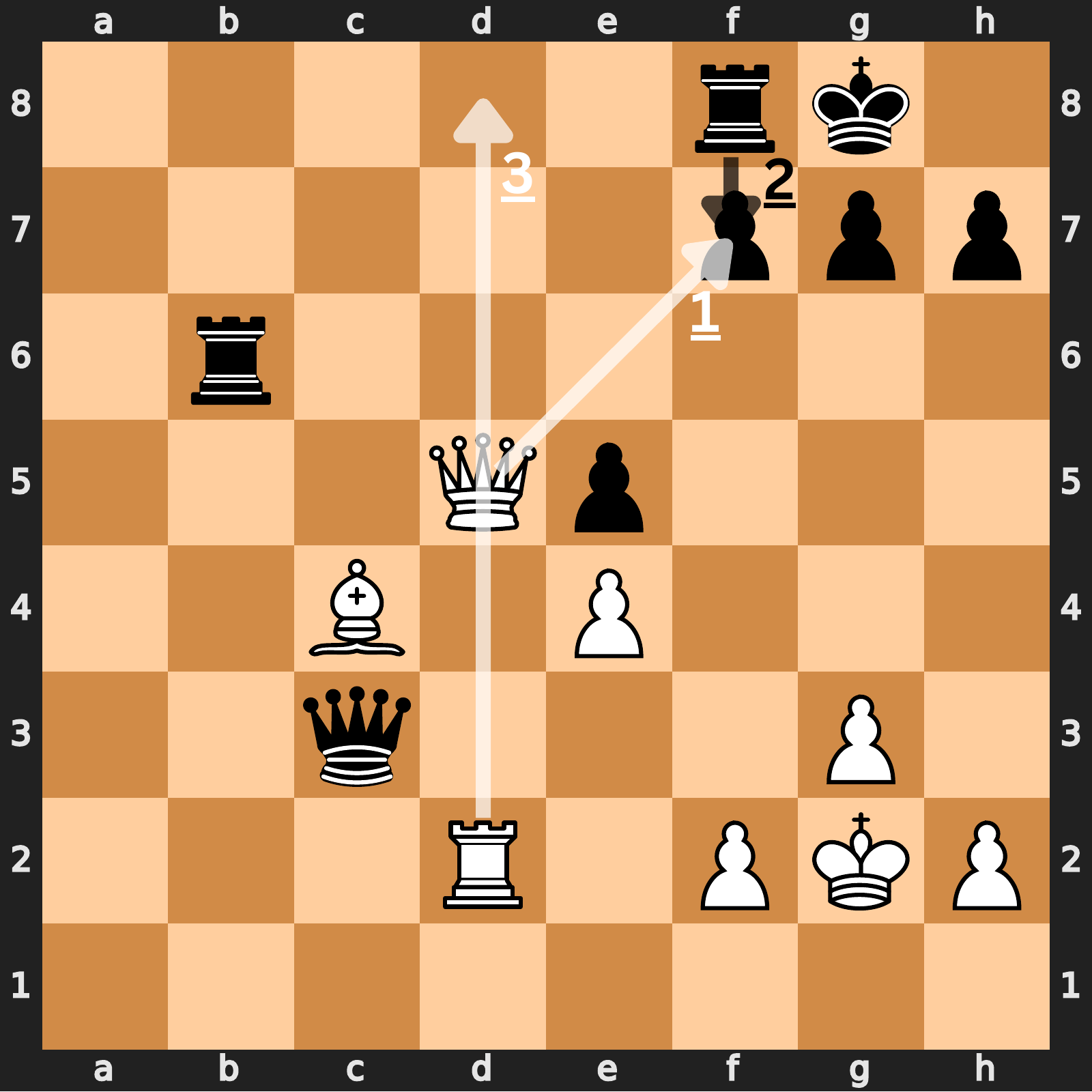}
    \caption{Pin puzzle}
    \label{rethink:fig:chess-pin-puzzle}
  \end{subfigure}
  \caption{Examples of tactical motifs used to fit and compose independently
  calibrated behavioral adaptations.}
  \label{rethink:fig:chess-environment}
\end{figure}

We use a frozen Maia-1500 policy, one of the Maia models trained to predict
human moves at a specified player rating~\cite{McIlroyYoung2020Maia}. We draw
puzzles from the open Lichess puzzle database~\cite{LichessPuzzles} and select
three tactical motifs: forks, pins, and skewers;
Figure~\ref{rethink:fig:chess-environment} shows illustrative fork and pin
puzzles. For each puzzle, the policy receives a binary reward of one when it
selects the canonical move and zero otherwise. For each motif, we construct an
independent adaptation from 40 training puzzles. Fit, calibration, and test
puzzles are disjoint. Each adaptation is calibrated independently to a fixed
action-KL budget. Compositions combine these isolated adaptations without an
additional composition-level rescaling, and we report the resulting composed
KL.

\paragraph{Canonical-move likelihood.}
We evaluate each adaptation on held-out puzzles using the negative
log-likelihood of the canonical move:
\begin{equation}
  \mathcal L_m(S)
  =
  -\mathbb E_{(o,a^\star)\sim\mathcal D_m}
  \left[\log \pi_S(a^\star\mid o)\right],
  \label{rethink:eq:chess-nll}
\end{equation}
where $S$ denotes the set of active adaptations.  Lower values indicate that
the policy assigns greater probability to the canonical move.  For motif $m$,
we define the gain from a composition $S$ as
$g_m(S)=\mathcal L_m(\varnothing)-\mathcal L_m(S)$.
\emph{Isolated gain} averages $g_m(\{m\})$ across motifs, while
\emph{composed gain} averages $g_m(S)$ with all three adaptations active.

\paragraph{Composition retention.}
To test whether composition sacrifices any individual objective, we report
\begin{equation}
  R_{\min}
  =
  \min_{\substack{|S|\geq2\\m\in S}}
  \frac{g_m(S)}{g_m(\{m\})}.
  \label{rethink:eq:chess-min-retention}
\end{equation}
A value of one means that every objective preserves its isolated improvement;
values above one indicate constructive interaction even for the least-retained
objective.

\paragraph{Intervention cost.}
We additionally report the mean action-distribution KL from the original policy
to the three-way composition and the number of scalars required to store the
three adaptations.  Further implementation and calibration details are
provided in the appendix.

\subsection{Results}

\paragraph{Comparison with adaptation baselines.}
Table~\ref{rethink:tab:chess-main-composition} compares all methods using the
same six residual blocks and behavioral-change budget.  PGS achieves the
strongest isolated gain and a composed gain comparable to LoRA and
Fine-tuning, while outperforming ReFT.  All methods preserve positive gains
for every motif after composition, showing that the three tactical objectives
are mutually compatible.  PGS obtains this performance with the smallest
stored adaptation, supporting its use as a compact mechanism for constructing
and composing behavioral interventions.

\begin{table}[H]
  \centering
  \scriptsize
  \setlength{\tabcolsep}{2.2pt}
  \adjustbox{max width=\linewidth}{%
  \begin{tabular}{lrrrrr}
    \toprule
    Method & Scalars $\downarrow$ & Isolated gain $\uparrow$ & Composed gain $\uparrow$ & Min.\ retention $\uparrow$ & Composed KL $\downarrow$ \\
    \midrule
    ReFT & $\underline{9{,}288}$ & $0.378 \pm 0.007$ & $0.761 \pm 0.012$ & $1.494 \pm 0.023$ & $\underline{0.661 \pm 0.014}$ \\
    LoRA & $46{,}080$ & $0.378 \pm 0.013$ & $0.803 \pm 0.023$ & $\underline{1.556 \pm 0.042}$ & $\mathbf{0.658 \pm 0.015}$ \\
    Fine-tuning & $663{,}552$ & $\underline{0.382 \pm 0.012}$ & $\mathbf{0.807 \pm 0.020}$ & $\mathbf{1.576 \pm 0.028}$ & $0.670 \pm 0.012$ \\
    \midrule
    PGS (ours) & $\mathbf{1{,}152}$ & $\mathbf{0.403 \pm 0.005}$ & $\underline{0.804 \pm 0.014}$ & $1.528 \pm 0.011$ & $0.689 \pm 0.019$ \\
    \bottomrule
  \end{tabular}%
  }
  \caption{Matched-support chess composition across all six residual blocks. Values are mean $\pm$ sample standard deviation across five fit seeds.}
  \label{rethink:tab:chess-main-composition}
\end{table}

\paragraph{Robustness across policy skill.}
We repeat the matched composition comparison with Maia-1100 and Maia-1900,
using the same protocol as for Maia-1500 and comparing every method against
its corresponding frozen base policy.  This experiment tests whether the
relative composition behavior is specific to one Maia checkpoint or persists
across policies trained to imitate different player skill levels.  Every
method improves all three checkpoints in isolation and after composition.
The leading baseline varies by checkpoint, while PGS remains close to the
strongest composed result and retains its substantially smaller stored
adaptation.  The composition effect is therefore not specific to Maia-1500,
but the experiment does not support a checkpoint-independent ranking of
adaptation methods.

\begin{table}[H]
  \centering
  \scriptsize
  \setlength{\tabcolsep}{2.2pt}
  \adjustbox{max width=\linewidth}{%
  \begin{tabular}{llrrrrr}
    \toprule
    Checkpoint & Method & Scalars $\downarrow$ & Isolated gain $\uparrow$ & Composed gain $\uparrow$ & Min.\ retention $\uparrow$ & Composed KL $\downarrow$ \\
    \midrule
    maia-1100 & ReFT & $\underline{9{,}288}$ & $\mathbf{0.449 \pm 0.005}$ & $0.919 \pm 0.012$ & $1.557 \pm 0.023$ & $0.661 \pm 0.012$ \\
    maia-1100 & LoRA & $46{,}080$ & $\underline{0.445 \pm 0.009}$ & $\underline{0.928 \pm 0.027}$ & $\underline{1.580 \pm 0.026}$ & $\mathbf{0.642 \pm 0.010}$ \\
    maia-1100 & Fine-tuning & $663{,}552$ & $0.444 \pm 0.008$ & $\mathbf{0.931 \pm 0.016}$ & $\mathbf{1.585 \pm 0.019}$ & $0.661 \pm 0.005$ \\
    \midrule
    maia-1100 & PGS (ours) & $\mathbf{1{,}152}$ & $0.425 \pm 0.008$ & $0.897 \pm 0.021$ & $1.570 \pm 0.029$ & $\underline{0.644 \pm 0.021}$ \\
    \midrule
    maia-1500 & ReFT & $\underline{9{,}288}$ & $0.378 \pm 0.007$ & $0.762 \pm 0.012$ & $1.495 \pm 0.022$ & $\underline{0.661 \pm 0.014}$ \\
    maia-1500 & LoRA & $46{,}080$ & $0.378 \pm 0.013$ & $0.804 \pm 0.023$ & $\underline{1.556 \pm 0.044}$ & $\mathbf{0.659 \pm 0.015}$ \\
    maia-1500 & Fine-tuning & $663{,}552$ & $\underline{0.382 \pm 0.012}$ & $\mathbf{0.807 \pm 0.020}$ & $\mathbf{1.576 \pm 0.028}$ & $0.670 \pm 0.012$ \\
    \midrule
    maia-1500 & PGS (ours) & $\mathbf{1{,}152}$ & $\mathbf{0.403 \pm 0.005}$ & $\underline{0.805 \pm 0.014}$ & $1.527 \pm 0.011$ & $0.690 \pm 0.019$ \\
    \midrule
    maia-1900 & ReFT & $\underline{9{,}288}$ & $\mathbf{0.428 \pm 0.003}$ & $\mathbf{0.933 \pm 0.010}$ & $\mathbf{1.608 \pm 0.064}$ & $\underline{0.690 \pm 0.027}$ \\
    maia-1900 & LoRA & $46{,}080$ & $0.414 \pm 0.005$ & $0.917 \pm 0.017$ & $1.567 \pm 0.031$ & $0.702 \pm 0.014$ \\
    maia-1900 & Fine-tuning & $663{,}552$ & $0.412 \pm 0.006$ & $0.911 \pm 0.021$ & $1.575 \pm 0.027$ & $0.706 \pm 0.016$ \\
    \midrule
    maia-1900 & PGS (ours) & $\mathbf{1{,}152}$ & $\underline{0.418 \pm 0.005}$ & $\underline{0.928 \pm 0.008}$ & $\underline{1.608 \pm 0.027}$ & $\mathbf{0.624 \pm 0.013}$ \\
    \bottomrule
  \end{tabular}%
  }
  \caption{Matched-support chess composition across Maia checkpoints. Values are mean $\pm$ sample standard deviation across five fit seeds.}
  \label{rethink:tab:chess-main-models}
\end{table}

\paragraph{Layer allocation.}
Table~\ref{rethink:tab:chess-cross-layer} compares three ways of allocating the
PGS adaptations across layers.  The same-layer condition places every motif at
the strongest common site according to its fit-side gradient norm.  The
gradient-selected condition assigns each motif to a distinct layer to maximize
the combined gradient-norm criterion.  The permuted condition uses the same
distinct layers but rotates their motif assignments.

The comparison reveals a trade-off between isolated strength and composition.
The common layer produces the strongest isolated adaptations, the
gradient-selected allocation gives the strongest minimum retention, and the
permuted allocation gives the strongest composed gain with the lowest policy
change.  Distributing adaptations across layers can therefore improve
composition efficiency, but isolated gradient strength does not identify the
best joint allocation.

\begin{table}[H]
  \centering
  \scriptsize
  \setlength{\tabcolsep}{2.2pt}
  \adjustbox{max width=\linewidth}{%
  \begin{tabular}{lrrrrr}
    \toprule
    Placement & Scalars $\downarrow$ & Isolated gain $\uparrow$ & Composed gain $\uparrow$ & Min.\ retention $\uparrow$ & Composed KL $\downarrow$ \\
    \midrule
    Same layer & $\mathbf{192}$ & $\mathbf{0.416 \pm 0.007}$ & $\underline{0.764 \pm 0.015}$ & $\underline{1.526 \pm 0.027}$ & $0.716 \pm 0.009$ \\
    Gradient-selected cross-layer & $\mathbf{192}$ & $0.356 \pm 0.012$ & $0.754 \pm 0.022$ & $\mathbf{1.600 \pm 0.032}$ & $\underline{0.644 \pm 0.021}$ \\
    Permuted cross-layer & $\mathbf{192}$ & $\underline{0.388 \pm 0.009}$ & $\mathbf{0.774 \pm 0.020}$ & $1.472 \pm 0.025$ & $\mathbf{0.615 \pm 0.025}$ \\
    \bottomrule
  \end{tabular}%
  }
  \caption{PGS composition with all objectives at one layer or distributed across distinct layers. Values are mean $\pm$ sample standard deviation across five fit seeds under the exact expected canonical-action objective.}
  \label{rethink:tab:chess-cross-layer}
\end{table}

\section{Policy Diversification in Football}
\label{rethink:sec:football-affinity}

\begin{figure}[ht]
  \centering
  \includegraphics[width=\linewidth]{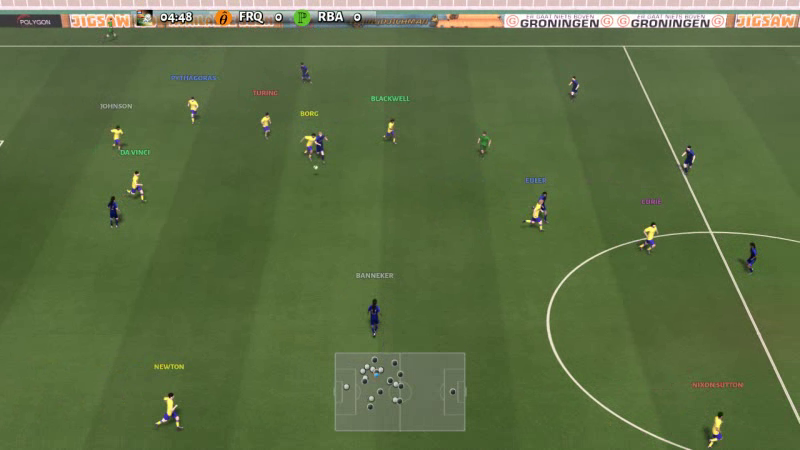}
  \caption{Google Research Football environment used for the behavior-steering
  study~\cite{Kurach2019GoogleRF}.}
  \label{rethink:fig:football-environment}
\end{figure}

\subsection{Experimental Setup}

We study frozen team policies released by
\citet{Song2024EmpiricalGRF}: \texttt{defensive\_passer},
\texttt{flank\_pass}, \texttt{group\_pressure}, and
\texttt{offensive\_passer}. These policies implement distinct learned football
strategies and serve as both controlled teams and opponents. We ask whether
PGS can produce temporary behavioral variants of these policies without
retraining them. Such variants could support more diverse populations for
cross-play evaluation, opponent curricula, or training
leagues~\cite{Bettini2024ControllingBD}.

We study three behavioral objectives: completed passes, possession regains,
and shot attempts. Each objective rewards its corresponding event, and we
evaluate it using the matched held-out event count and behavior rate. The
appendix protocol records the other configurable football objectives, which are
outside the main study.

\begin{table}[H]
  \centering
  \scriptsize
  \adjustbox{max width=\linewidth}{%
  \begin{tabular}{lrrr}
    \toprule
    Self-play policy & Completed passes $\uparrow$ & Possession regains $\uparrow$ & Shot attempts $\uparrow$ \\
    \midrule
    \texttt{defensive\_passer} & \textbf{116.00 $\pm$ 28.04} & 26.00 $\pm$ 2.24 & 7.20 $\pm$ 4.71 \\
    \texttt{flank\_pass} & \underline{70.20 $\pm$ 14.79} & 35.60 $\pm$ 3.97 & 9.40 $\pm$ 7.16 \\
    \texttt{group\_pressure} & 6.00 $\pm$ 4.58 & \textbf{91.60 $\pm$ 13.16} & \underline{10.60 $\pm$ 8.56} \\
    \texttt{offensive\_passer} & 49.20 $\pm$ 21.51 & \underline{40.80 $\pm$ 4.38} & \textbf{23.60 $\pm$ 11.95} \\
    \bottomrule
  \end{tabular}%
  }
  \caption{Unsteered full-match main-study behaviors; values are mean $\pm$ sample standard deviation over five self-play matches.}
  \label{tab:football-main-behavior-description}
\end{table}

\subsection{Results}

\paragraph{Passing Behavior.}
We use completed passing as the target behavior for the main comparison
between PGS, ReFT, LoRA, Fine-tuning, and a PGS-norm random-activation
baseline. All methods use full-match self-play trajectories and are calibrated to the same action-KL budget.
Table~\ref{tab:football-main-pass} reports full-match outcomes for the
unmodified \texttt{group\_pressure} policy and each intervention. PGS produces
a pronounced shift toward passing, yielding the highest mean pass-attempt
count and a completed-pass count comparable to the strongest learned
adaptation. In contrast, random activation and Fine-tuning remain near the
unmodified policy, showing that neither intervention magnitude nor direct
parameter optimization is sufficient to induce the target behavior at the
same policy-change budget. Variation across held-out matches nevertheless
remains substantial.

\begin{table}[H]
  \centering
  \small
  \setlength{\tabcolsep}{4.5pt}
  \adjustbox{max width=\columnwidth}{%
  \begin{tabular}{lrrrr}
    \toprule
    Method & Attempts $\uparrow$ & Completed $\uparrow$ & Goal diff. $\uparrow$ & \multicolumn{1}{c}{KL $\downarrow$} \\
    \midrule
    Base & \ensuremath{12.80 \pm 7.73} & \ensuremath{6.00 \pm 4.58} & \ensuremath{0.20 \pm 0.45} & -- \\
    Random activation & \ensuremath{17.80 \pm 11.52} & \ensuremath{5.60 \pm 6.58} & \ensuremath{\underline{-0.20 \pm 0.84}} & \ensuremath{0.11 \pm 0.01} \\
    ReFT & \ensuremath{26.00 \pm 13.06} & \ensuremath{16.00 \pm 13.51} & \ensuremath{\mathbf{0.00 \pm 0.00}} & \ensuremath{\mathbf{0.10 \pm 0.01}} \\
    LoRA & \ensuremath{\underline{56.40 \pm 23.77}} & \ensuremath{\mathbf{32.00 \pm 17.68}} & \ensuremath{-0.40 \pm 0.55} & \ensuremath{0.11 \pm 0.03} \\
    Fine-tuning & \ensuremath{10.00 \pm 11.90} & \ensuremath{4.80 \pm 6.53} & \ensuremath{\mathbf{0.00 \pm 0.00}} & \ensuremath{\underline{0.10 \pm 0.03}} \\
    \midrule
    PGS (ours) & \ensuremath{\mathbf{64.00 \pm 30.98}} & \ensuremath{\underline{29.80 \pm 16.04}} & \ensuremath{-0.40 \pm 1.14} & \ensuremath{0.12 \pm 0.01} \\
    \bottomrule
  \end{tabular}%
  }
  \caption{Full-match outcomes for shared interventions to \texttt{group\_pressure} in self-play. Values are mean $\pm$ sample standard deviation over held-out match seeds.}
  \label{tab:football-main-pass}
\end{table}

PGS interventions also transfer across opponents. We fit one intervention from
\texttt{group\_pressure} matches against each non-self opponent, then evaluate
each frozen intervention against all four policies. Each intervention increases
mean completed passes relative to the unmodified policy across evaluation
opponents, while the intervention fitted from trajectories with greater passing
exposure is consistently strongest. Thus, PGS can encode a transferable team
behavior, with its effectiveness shaped by the fitting trajectories.

\begin{table}[H]
  \centering
  \small
  \setlength{\tabcolsep}{4.5pt}
  \adjustbox{max width=\columnwidth}{%
  \begin{tabular}{lrrr|r}
    \toprule
    Fit opponent & defensive\_passer & flank\_pass & offensive\_passer & group\_pressure \\
    \midrule
    defensive\_passer & \ensuremath{36.20 \pm 10.59} & \ensuremath{39.00 \pm 23.27} & \ensuremath{29.80 \pm 14.79} & \ensuremath{9.00 \pm 6.04} \\
    flank\_pass & \ensuremath{47.40 \pm 22.57} & \ensuremath{46.40 \pm 24.19} & \ensuremath{40.00 \pm 14.40} & \ensuremath{22.60 \pm 11.22} \\
    offensive\_passer & \ensuremath{16.80 \pm 7.98} & \ensuremath{29.40 \pm 18.46} & \ensuremath{28.20 \pm 12.87} & \ensuremath{10.80 \pm 7.40} \\
    \bottomrule
  \end{tabular}%
  }
  \caption{Cross-play completed passes from PGS applied to \texttt{group\_pressure} (mean $\pm$ sample standard deviation over held-out match seeds).}
  \label{tab:football-main-pass-cross-play}
\end{table}

\paragraph{Behavioral Diversity and Fitting Sensitivity.}
Beyond the main passing comparison, we apply PGS across multiple team policies
and behavioral objectives. We compare directions fitted from each target
policy's demonstrations
(Table~\ref{tab:football-main-self-play-pgs-outcomes}) with directions fitted
from the policy strongest in the target behavior
(Table~\ref{tab:football-main-self-play-offline-pgs-outcomes}). Neither fitting
source dominates uniformly, showing that steering effectiveness depends on the
interaction between the controlled policy, behavioral objective, and fitting
trajectories. This dependence is most pronounced for possession regains, a
team-level event requiring agent-level credit assignment. These results extend
PGS beyond a single passing intervention while identifying fit-data selection
as a central challenge for reliable multi-agent steering.

\begin{table}[H]
  \centering
  \small
  \setlength{\tabcolsep}{4.5pt}
  \adjustbox{max width=\columnwidth}{%
  \begin{tabular}{lrrr}
    \toprule
    Self-play policy & Completed passes & Possession regains & Shot attempts \\
    \midrule
    defensive\_passer & \ensuremath{133.60 \pm 70.26} & \ensuremath{25.60 \pm 5.13} & \ensuremath{5.60 \pm 4.34} \\
    flank\_pass & \ensuremath{28.20 \pm 14.25} & \ensuremath{35.00 \pm 8.60} & \ensuremath{13.40 \pm 14.15} \\
    group\_pressure & \ensuremath{12.00 \pm 9.95} & \ensuremath{86.20 \pm 8.70} & \ensuremath{9.40 \pm 14.62} \\
    offensive\_passer & \ensuremath{64.40 \pm 24.28} & \ensuremath{42.60 \pm 7.70} & \ensuremath{15.20 \pm 7.29} \\
    \bottomrule
  \end{tabular}%
  }
  \caption{Absolute self-play outcomes from PGS with target-policy demonstrations from both sides (mean $\pm$ sample standard deviation over five held-out match seeds).}
  \label{tab:football-main-self-play-pgs-outcomes}
\end{table}

\begin{table}[H]
  \centering
  \small
  \setlength{\tabcolsep}{4.5pt}
  \adjustbox{max width=\columnwidth}{%
  \begin{tabular}{lrrr}
    \toprule
    Self-play policy & Completed passes & Possession regains & Shot attempts \\
    \midrule
    defensive\_passer & -- & \ensuremath{26.60 \pm 1.14} & \ensuremath{13.80 \pm 6.02} \\
    flank\_pass & \ensuremath{53.00 \pm 16.29} & \ensuremath{31.60 \pm 3.71} & \ensuremath{8.80 \pm 7.33} \\
    group\_pressure & \ensuremath{7.80 \pm 5.31} & -- & \ensuremath{8.20 \pm 5.76} \\
    offensive\_passer & \ensuremath{49.00 \pm 15.75} & \ensuremath{43.20 \pm 6.22} & -- \\
    \bottomrule
  \end{tabular}%
  }
  \caption{Absolute self-play outcomes from PGS with offline source-policy demonstrations from both sides (mean $\pm$ sample standard deviation over five held-out match seeds).}
  \label{tab:football-main-self-play-offline-pgs-outcomes}
\end{table}

\section{Discussion}
\label{rethink:sec:discussion}

\paragraph{What PGS offers.}
PGS turns scalar behavioral feedback into a removable activation intervention by
accumulating return-weighted action-score gradients from a fixed batch. Its
advantage is operational: it constructs a direction in one gradient estimation
step, leaves the base-policy parameters unchanged, and permits the same stored
vector to be strengthened, weakened, or removed at inference time. PGS is
therefore suited to settings where low-step, low-storage, post-training control
is valuable, while iterative fine-tuning, LoRA, and ReFT provide complementary
routes for optimizing a selected behavioral objective.

\paragraph{Reusable behavioral interventions.}
Chess and football expose complementary forms of reuse. In chess, independently
constructed PGS vectors retain their tactical effects under composition,
although effective activation sites need not compose equally well. In football,
a fitted passing intervention transfers across opponents without refitting,
although its effectiveness depends on the fitting trajectories. Together, these
results motivate libraries of reusable behavioral controls accompanied by
composition and data-selection rules.

\paragraph{Evaluation boundaries.}
PGS is a local adaptation method: Fisher scaling is a local Fisher
approximation, activation-site choice remains consequential, and a behavioral
effect does not by itself establish a semantic or mechanistic interpretation of
the fitted vector. Evaluations should therefore report behavioral gain together
with achieved policy KL, task outcomes, seed-to-seed variation, storage, and
inference cost. Chess puzzle likelihood measures tactical preference rather
than playing strength, while football target-behavior gains must be read
alongside their collateral effects and competitive-match outcomes.

\paragraph{Future work.}
PGS suggests three extensions. Off-policy fitting could use cumulative or
marginalized importance weights, doubly robust policy-gradient
estimators~\cite{Huang2020DoublyRobustPG}, or temporal-difference gradient
estimation~\cite{Tosatto2022TDPolicyGradient}. When the local Fisher
approximation is inaccurate at finite scale, empirical action-KL backtracking
could refine the closed-form coefficient, paralleling trust-region
optimization~\cite{Schulman2015TRPO} while calibrating a fixed activation-space
direction. Finally, natural-gradient preconditioning with
$F^{-1}v_{\mathrm{PGS}}$~\cite{Kakade2001Natural} could alter the direction
rather than merely scale it.

\section{Related Work}
\label{rethink:sec:related-work}

\subsection{Post-hoc Steering and Adaptation}

Preference Goal Tuning (PGT) keeps a goal-conditioned policy frozen and
optimizes only its continuous goal embedding from trajectory preferences
\cite{Zhao2026PGT}. Like PGS, it treats post-training adaptation as
inference-time control rather than parameter fine-tuning. Unlike PGS, PGT
requires a policy with a native goal-conditioning interface and iteratively
fits its control variable, whereas PGS constructs a removable activation vector
from scalar behavioral returns for policies with no such interface.
Policy Gradient Guidance instead trains conditional and unconditional policy
branches to expose a test-time guidance scale~\cite{Qi2025PolicyGradientGuidance};
PGS constructs its intervention after the base policy has been trained.

Activation-steering methods derive interventions from contrastive examples,
learned attributes, or rollout outcomes
\cite{zou2023representation,Li2023InferenceTimeIntervention,
Turner2023ActivationEngineering,rimsky-etal-2024-steering,
stoehr-etal-2024-activation,oozeer2025ksteering,miao2026coast}.
PPLM uses gradients from an attribute model to perturb hidden states iteratively
during decoding, while LatentSeek applies policy gradients to per-instance
latent sequences at test time~\cite{Dathathri2020PPLM,Li2025LatentSeek}; PGS
instead aggregates return-weighted action-score gradients into a reusable,
state-independent vector. ReFT and LoRA are related learned adaptation mechanisms
\cite{hu2021loralowrankadaptationlarge,wu2024reft}; we use them as baselines
rather than claiming their parameterizations as contributions.

\subsection{Task Vectors and Composition}
Task arithmetic composes fine-tuning deltas, and early gradient steps can
themselves form task vectors
\cite{Ilharco2022EditingMW,zhou2025taskvectorsgradients}. AdapterFusion instead
learns to combine separately trained task adapters
\cite{Pfeiffer2021AdapterFusion}. Activation-space composition has likewise
been studied through dynamic multi-property steering and compositional
steering tokens
\cite{Scalena2024DynamicComposition,Radevski2026CompositionalSteering}.
Preference Vectors use DPO to train models for preferred and label-reversed
objectives, then subtract their weights to form scalable task vectors
\cite{Rafailov2023DPO,Liang2026PreferenceVectors}. Contrastive weight steering
similarly constructs parameter-space directions from opposing behaviors
\cite{Fierro2025WeightArithmetic}.
PGS does not claim novelty for the correspondence between gradients and task
vectors. Its focus is constructing removable activation-space vectors from
scalar action or trajectory returns and calibrating their individual and
composed effects in policy space.

\subsection{Multi-agent Adaptation}
Explicit opponent models can learn opponent strategy patterns
\cite{He2016OpponentModeling}; hierarchical opponent models can infer latent
goals and support few-shot adaptation to unseen policies
\cite{Huang2024OpponentAdaptation}. Our football study does not infer an explicit
opponent model; it tests whether interventions fitted under one strategic
context transfer or specialize across agent roles and opponents. Shared-policy
MARL makes role specificity important: agents commonly act through shared
policy parameters~\cite{yu2022surprising}, yet representing distinct agent
policies or tasks requires agent-identifying information
\cite{terry2023revisitingparametersharingmultiagent}.

\section{Conclusion}
\label{rethink:sec:conclusion}

We introduced Policy Gradient Steering, which turns temporary behavioral
objectives into compact, removable activation interventions calibrated under a
policy-KL budget. Across controlled route choice and compositional chess
steering, our results show that policy-gradient credit can support
inference-time behavioral adaptation without modifying the base policy; the
football study extends this question to multi-agent contexts. PGS therefore
provides a practical interface for introducing, combining, and removing
behavioral preferences after training.

\clearpage
\bibliography{main}

\clearpage
\appendix
\twocolumn[
  \begin{center}
    {\LARGE\bfseries Appendix\par}
    \vspace{0.5em}
  \end{center}
]

\section{Gridworld Additional Details}
\label{rethink:app:gridworld}

\subsection{Protocol}

\begin{itemize}
  \item Freeze the policy before data collection and use disjoint fit,
  validation, and evaluation streams.
  \item Apply activation methods at declared post-ReLU sites, rank-one LoRA to
  all linear weights, and fit the optimized realizations with fixed-batch
  updates.
  \item Fit PGS for one update and the optimized parameter-space
  realizations for three updates, using learning rate $10^{-2}$. Fit PGS
  jointly at \texttt{h1\_post} and
  \texttt{h2\_post}, then normalize the concatenated activation-offset vector
  to unit Euclidean norm. Leave other optimized tensors unchanged and apply
  them at scale $\alpha/10^{-2}$.
  \item In the primary contrastive comparisons, treat $\alpha$ as application
  strength and select it on validation data only. In the parameterization
  comparison, instead compute $\alpha$ from directional Fisher curvature for
  a target action KL of $0.1$, without a behavioral scale sweep.
  \item Evaluate the selected object in both directions on the same held-out
  stream and report unsuccessful routes explicitly.
\end{itemize}

The successful-route preference is
\begin{equation}
  R_{\mathrm{route}}(\tau)=
  \begin{cases}
    +1,&\text{top-route success},\\
    -1,&\text{bottom-route success},\\
    r_{\mathrm{fail}},&\text{failure}.
  \end{cases}
  \label{rethink:eq:gridworld-reward}
\end{equation}

\subsection{Change Inspection}

For both primary fit regimes, we reconstruct each seed-specific
validation-selected PGS intervention for all five fit seeds, evaluate every action
at every nonterminal state, and average probabilities across fitted seeds. The
plots expose changes hidden by route-level aggregates without selecting a
favorable fitted direction.

\begin{figure}[ht]
  \centering
  \includegraphics[width=\linewidth]{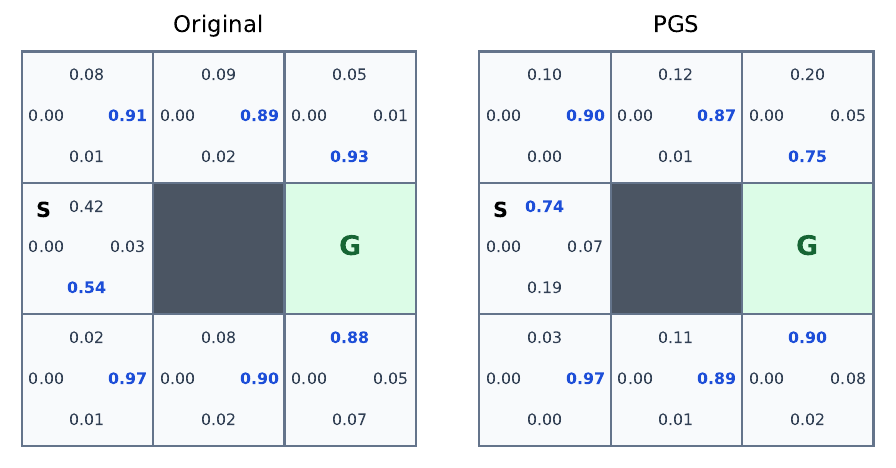}
  \caption{Mean statewise action probabilities for the original policy and
  seed-specific validation-selected PGS fitted from 20 frozen-policy rollouts.
  Each row compares the original policy (left) with PGS (right). Blue bold
  text marks each panel's greedy action; intervention probabilities are
  averaged over five fitted seeds.}
  \label{rethink:fig:gridworld-change-inspection-rollouts}
\end{figure}

\begin{figure}[ht]
  \centering
  \includegraphics[width=\linewidth]{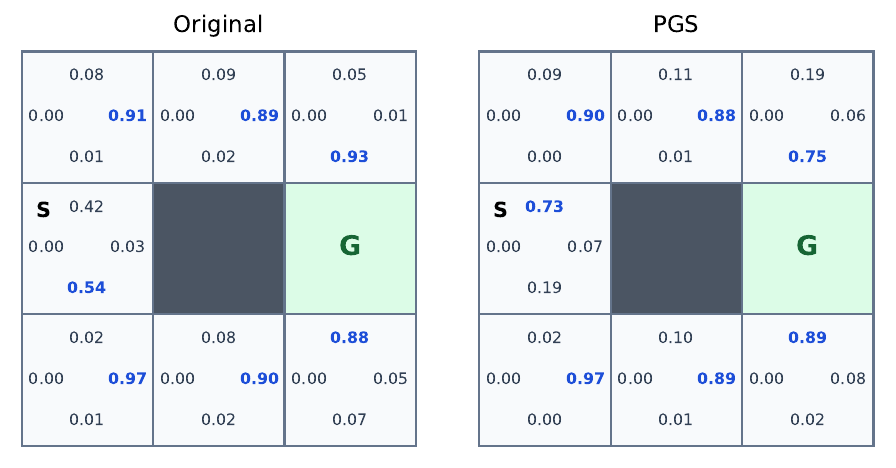}
  \caption{Mean statewise action probabilities for the original policy and
  seed-specific validation-selected PGS fitted from one demonstration per
  route. Each row compares the original policy (left) with PGS (right). Blue
  bold text marks each panel's greedy action; intervention probabilities are
  averaged over five fitted seeds.}
  \label{rethink:fig:gridworld-change-inspection-demonstrations}
\end{figure}
\FloatBarrier

\subsection{Ablations}

All ablations reuse the rollout protocol and paired validation and evaluation
streams. Except for the learning-rate sweep, each setting is selected on
validation data before held-out evaluation.

\paragraph{Fit trajectory count.}
PGS produces a directional change even from the smallest fit batch, but its
effect does not improve monotonically with additional trajectories. Neither
PGS nor fine-tuning exhibits a uniform sample-scaling trend over these small
fit batches.

\begin{table}[H]
  \centering
  \small
  \setlength{\tabcolsep}{4.5pt}
  \adjustbox{max width=\columnwidth}{%
  \begin{tabular}{lcc}
    \toprule
    Fit trajectories & PGS (ours) & Fine-tuning \\
    \midrule
    4 & $0.78 \pm 0.15$ & $0.83 \pm 0.18$ \\
    8 & $0.71 \pm 0.05$ & $0.70 \pm 0.10$ \\
    16 & $\mathbf{0.86 \pm 0.04}$ & $\underline{0.91 \pm 0.06}$ \\
    20 & $\underline{0.86 \pm 0.03}$ & $\mathbf{0.92 \pm 0.04}$ \\
    32 & $0.72 \pm 0.05$ & $0.74 \pm 0.08$ \\
    64 & $0.85 \pm 0.02$ & $0.73 \pm 0.06$ \\
    \bottomrule
  \end{tabular}%
  }
  \caption{Gridworld fit trajectories ablation. Cells report mean $\pm$ sample standard deviation of disjoint test top-route probability over five fit seeds.}
  \label{tab:gridworld-fit-budget}
\end{table}

\paragraph{Return treatment.}
We compare raw returns, batch-centered returns, standardized returns, and a
fitted linear state baseline. After normalizing the resulting steering
direction, standardization is equivalent to centering, while the fitted
baseline provides no clear behavioral improvement. We therefore use the simple
batch-centered return in the main experiments.

\begin{table}[H]
  \centering
  \small
  \setlength{\tabcolsep}{4.5pt}
  \adjustbox{max width=\columnwidth}{%
  \begin{tabular}{lccc}
    \toprule
    Return treatment & Selected $\alpha$ & Top route $\uparrow$ & KL $\downarrow$ \\
    \midrule
    none & $0.42$ & $\mathbf{0.82 \pm 0.05}$ & $\underline{0.10733}$ \\
    centered & $0.42$ & $\underline{0.82 \pm 0.07}$ & $0.10745$ \\
    standardized & $0.42$ & $\underline{0.82 \pm 0.07}$ & $0.10745$ \\
    linear baseline & $0.42$ & $0.81 \pm 0.09$ & $\mathbf{0.10729}$ \\
    \bottomrule
  \end{tabular}%
  }
  \caption{Gridworld return treatment and fitted-baseline comparison.}
  \label{tab:gridworld-normalization}
\end{table}

\paragraph{Intervention site.}
PGS is sensitive to where the offset is applied. Intervening at the first
hidden layer is more effective than at the default second-layer site, while
jointly fitting offsets at both post-ReLU sites produces the strongest route
preference, with the two pre-ReLU sites a close second. We therefore use the
joint post-ReLU sites in the main comparison.

\begin{table}[H]
  \centering
  \small
  \setlength{\tabcolsep}{4.5pt}
  \adjustbox{max width=\columnwidth}{%
  \begin{tabular}{lccc}
    \toprule
    Intervention sites & Best $\alpha$ & Top route prob. $\uparrow$ & Path length $\downarrow$ \\
    \midrule
    \texttt{h1\_pre} & 0.5 & $0.78 \pm 0.02$ & $\underline{4.50 \pm 0.10}$ \\
    \texttt{h1\_post} & 0.5 & $0.81 \pm 0.02$ & $4.57 \pm 0.07$ \\
    \texttt{h2\_pre} & 0.75 & $0.79 \pm 0.01$ & $\mathbf{4.49 \pm 0.07}$ \\
    \texttt{h2\_post} & 0.75 & $0.81 \pm 0.03$ & $4.55 \pm 0.09$ \\
    \texttt{h1\_pre+h2\_pre} & 0.5 & $\underline{0.85 \pm 0.03}$ & $4.56 \pm 0.12$ \\
    \texttt{h1\_post+h2\_post} & 0.5 & $\mathbf{0.86 \pm 0.03}$ & $4.58 \pm 0.12$ \\
    \bottomrule
  \end{tabular}%
  }
  \caption{Gridworld intervention sites ablation using policy rollouts. Alpha is selected separately for each setting on validation data; values report mean $\pm$ sample standard deviation on the disjoint test stream over five fit seeds.}
  \label{tab:gridworld-site-topology}
\end{table}

\paragraph{Optimizer strength.}
After three fixed-batch updates, unit-normalized activation offsets are stable
from learning rates $10^{-4}$ through $10^{-2}$ and weaken at larger rates.
Full-policy deltas divided by their learning rate are stable through $10^{-1}$
but degrade sharply at learning rate $1$. We fix the shared optimizer learning
rate to $10^{-2}$; application strength remains method-specific and is selected
separately.
\begin{table}[H]
  \centering
  \small
  \setlength{\tabcolsep}{4.5pt}
  \adjustbox{max width=\columnwidth}{%
  \begin{tabular}{lcc}
    \toprule
    Optimizer LR & PGS (ours) & Fine-tuning \\
    \midrule
    $10^{-4}$ & $0.5,\ \mathbf{0.86 \pm 0.03}$ & $0.25,\ \underline{0.92 \pm 0.05}$ \\
    $10^{-3}$ & $0.5,\ \mathbf{0.86 \pm 0.03}$ & $0.25,\ 0.92 \pm 0.05$ \\
    $10^{-2}$ & $0.5,\ \mathbf{0.86 \pm 0.03}$ & $0.25,\ \mathbf{0.92 \pm 0.04}$ \\
    $10^{-1}$ & $0.3,\ \underline{0.72 \pm 0.04}$ & $0.25,\ \underline{0.92 \pm 0.05}$ \\
    $10^{0}$ & $0.3,\ 0.69 \pm 0.03$ & $0.25,\ 0.70 \pm 0.14$ \\
    \bottomrule
  \end{tabular}%
  }
  \caption{Gridworld optimizer-strength ablation. Cells report the validation-selected $\alpha$ and disjoint-test top-route probability over five fit seeds.}
  \label{tab:gridworld-optim-strength}
\end{table}

\paragraph{Optimization steps.}
After unit normalization, the PGS direction is behaviorally stable from one to
five fixed-batch updates and changes little at ten. Fine-tuning is strongest
after three updates and degrades with further reuse of the same batch. We
therefore use one update for PGS and three for fine-tuning.

\begin{table}[H]
  \centering
  \small
  \setlength{\tabcolsep}{4.5pt}
  \adjustbox{max width=\columnwidth}{%
  \begin{tabular}{l@{\hspace{10pt}}c@{\hspace{10pt}}c}
    \toprule
    Updates & PGS (ours) & Fine-tuning \\
    \midrule
    1 & $\mathbf{0.86 \pm 0.03}$ & $0.87 \pm 0.08$ \\
    2 & $\mathbf{0.86 \pm 0.03}$ & $\underline{0.90 \pm 0.05}$ \\
    3 & $\mathbf{0.86 \pm 0.03}$ & $\mathbf{0.92 \pm 0.04}$ \\
    5 & $\mathbf{0.86 \pm 0.03}$ & $0.87 \pm 0.07$ \\
    10 & $\underline{0.85 \pm 0.03}$ & $0.87 \pm 0.07$ \\
    \bottomrule
  \end{tabular}%
  }
  \caption{Gridworld updates ablation. Cells report mean $\pm$ sample standard deviation of disjoint test top-route probability over five fit seeds.}
  \label{tab:gridworld-update-count}
\end{table}

\section{Additional Chess Details}
\label{rethink:app:chess}

\subsection{Encoding and Legal Actions}

LCZeroLens uses LCZero's representation and represents each position as a
current-player-oriented tensor with 112 planes of size
$8\times8$~\cite{The_LCZero_Authors_LeelaChessZero}. The first 104 planes
encode up to eight
positions of history using twelve piece planes and one repetition plane per
position; the remaining planes encode castling rights, side-to-move
orientation, the half-move clock, and constant metadata.  Positions with
Black to move are flipped into the same canonical learner orientation.

The policy head uses LCZero's 1,858-entry move encoding. At every decision we
mask logits for illegal moves and normalize the policy only over the legal set.
Dataset moves are stored in UCI notation and encoded relative to the current
board orientation, with decoded promotions matched to their legal moves.
Fitting, validation, and evaluation use the same representation, orientation
convention, and legal-action mask.

\subsection{Protocol}

\begin{itemize}
  \item Partition puzzle identities globally into exclusive fork, pin, and
  skewer fit, validation, and evaluation cohorts.
  \item Optimize the exact expected binary reward of the canonical legal
  action at each fixed offline puzzle state.
  \item Compare channel PGS with rank-four ReFT, rank-four LoRA, and
  Fine-tuning over all six residual blocks.  PGS and ReFT act after each
  block's second convolution and ReLU, while LoRA and Fine-tuning update the
  corresponding six convolutional weight tensors.
  \item Use one fitting update for PGS and three for ReFT, LoRA, and
  Fine-tuning.
  \item Estimate directional curvature from the realized action KL at
  coefficients $\pm 10^{-3}$ and set each independently fitted object's
  coefficient to $\alpha=\sqrt{2(0.1)/(v^\top Fv)}$.
  \item Add parameter deltas, apply ReFT residual maps in parallel, and sum PGS
  interventions when composing objectives. The main composition does not apply
  an additional joint-KL rescaling.
\end{itemize}

\subsection{Ablations}

All reported values are mean $\pm$ sample standard deviation across five fit
seeds.  Every condition uses the same disjoint calibration and test cohorts.

\paragraph{Layer support.}
We compare nested bottom-up and top-down residual support, a policy-head
control, and all six residual blocks.  Each independently fitted object
receives the same total isolated action-KL budget, so adding sites does not
increase the allowed policy change.  Bottom-up support is consistently more
effective than matched top-down support.  PGS achieves its strongest composed
gain with all six blocks, while the parameter-update baselines saturate with
four bottom-up blocks.  Head-only adaptation is weak for every method and
produces negative composed gain for PGS, indicating that the promoted result
depends on residual features rather than only the policy head.

\begin{table}[H]
  \centering
  \scriptsize
  \setlength{\tabcolsep}{2.2pt}
  \adjustbox{max width=\linewidth}{%
  \begin{tabular}{llrrrrr}
    \toprule
    Method & Support & Scalars $\downarrow$ & Isolated gain $\uparrow$ & Composed gain $\uparrow$ & Min.\ retention $\uparrow$ & Composed KL $\downarrow$ \\
    \midrule
    ReFT & Bottom-up 1 & $\mathbf{1{,}548}$ & $\mathbf{0.398 \pm 0.006}$ & $\underline{0.760 \pm 0.019}$ & $\mathbf{1.547 \pm 0.060}$ & $0.716 \pm 0.023$ \\
    ReFT & Bottom-up 2 & $\underline{3{,}096}$ & $0.386 \pm 0.005$ & $0.745 \pm 0.021$ & $\underline{1.540 \pm 0.056}$ & $0.703 \pm 0.031$ \\
    ReFT & Bottom-up 4 & $6{,}192$ & $\underline{0.393 \pm 0.006}$ & $\mathbf{0.767 \pm 0.018}$ & $1.537 \pm 0.022$ & $0.660 \pm 0.013$ \\
    ReFT & Top-down 1 & $\mathbf{1{,}548}$ & $0.158 \pm 0.015$ & $0.200 \pm 0.026$ & $0.902 \pm 0.060$ & $\underline{0.601 \pm 0.036}$ \\
    ReFT & Top-down 2 & $\underline{3{,}096}$ & $0.213 \pm 0.008$ & $0.364 \pm 0.019$ & $1.261 \pm 0.067$ & $0.623 \pm 0.051$ \\
    ReFT & Top-down 4 & $6{,}192$ & $0.325 \pm 0.010$ & $0.656 \pm 0.022$ & $1.458 \pm 0.062$ & $0.616 \pm 0.025$ \\
    ReFT & Head & $\mathbf{1{,}548}$ & $0.108 \pm 0.005$ & $0.006 \pm 0.054$ & $-0.559 \pm 0.884$ & $\mathbf{0.580 \pm 0.039}$ \\
    ReFT & All 6 & $9{,}288$ & $0.379 \pm 0.004$ & $0.759 \pm 0.006$ & $1.489 \pm 0.023$ & $0.667 \pm 0.012$ \\
    \midrule
    LoRA & Bottom-up 1 & $\mathbf{7{,}680}$ & $0.343 \pm 0.018$ & $0.754 \pm 0.011$ & $1.559 \pm 0.038$ & $\mathbf{0.528 \pm 0.013}$ \\
    LoRA & Bottom-up 2 & $\underline{15{,}360}$ & $0.355 \pm 0.014$ & $0.793 \pm 0.017$ & $\mathbf{1.594 \pm 0.074}$ & $0.629 \pm 0.027$ \\
    LoRA & Bottom-up 4 & $30{,}720$ & $\mathbf{0.379 \pm 0.009}$ & $\mathbf{0.800 \pm 0.032}$ & $\underline{1.584 \pm 0.041}$ & $0.657 \pm 0.022$ \\
    LoRA & Top-down 1 & $\mathbf{7{,}680}$ & $0.194 \pm 0.018$ & $0.340 \pm 0.032$ & $1.239 \pm 0.092$ & $\underline{0.540 \pm 0.020}$ \\
    LoRA & Top-down 2 & $\underline{15{,}360}$ & $0.273 \pm 0.005$ & $0.521 \pm 0.014$ & $1.402 \pm 0.089$ & $0.559 \pm 0.042$ \\
    LoRA & Top-down 4 & $30{,}720$ & $0.367 \pm 0.017$ & $0.713 \pm 0.039$ & $1.520 \pm 0.027$ & $0.599 \pm 0.042$ \\
    LoRA & Head & $15{,}552$ & $0.182 \pm 0.018$ & $0.196 \pm 0.033$ & $0.582 \pm 0.309$ & $0.613 \pm 0.018$ \\
    LoRA & All 6 & $46{,}080$ & $\underline{0.377 \pm 0.010}$ & $\underline{0.799 \pm 0.023}$ & $1.581 \pm 0.056$ & $0.654 \pm 0.014$ \\
    \midrule
    Fine-tuning & Bottom-up 1 & $\mathbf{110{,}592}$ & $0.351 \pm 0.015$ & $0.769 \pm 0.009$ & $\underline{1.585 \pm 0.046}$ & $\mathbf{0.533 \pm 0.015}$ \\
    Fine-tuning & Bottom-up 2 & $\underline{221{,}184}$ & $0.368 \pm 0.013$ & $\underline{0.812 \pm 0.013}$ & $\mathbf{1.612 \pm 0.041}$ & $0.658 \pm 0.014$ \\
    Fine-tuning & Bottom-up 4 & $442{,}368$ & $\mathbf{0.384 \pm 0.011}$ & $\mathbf{0.815 \pm 0.017}$ & $1.585 \pm 0.034$ & $0.663 \pm 0.014$ \\
    Fine-tuning & Top-down 1 & $\mathbf{110{,}592}$ & $0.233 \pm 0.005$ & $0.400 \pm 0.005$ & $1.330 \pm 0.030$ & $\underline{0.574 \pm 0.012}$ \\
    Fine-tuning & Top-down 2 & $\underline{221{,}184}$ & $0.290 \pm 0.011$ & $0.554 \pm 0.020$ & $1.505 \pm 0.017$ & $0.609 \pm 0.024$ \\
    Fine-tuning & Top-down 4 & $442{,}368$ & $0.369 \pm 0.007$ & $0.710 \pm 0.017$ & $1.532 \pm 0.020$ & $0.625 \pm 0.017$ \\
    Fine-tuning & Head & $248{,}832$ & $0.187 \pm 0.017$ & $0.223 \pm 0.028$ & $0.733 \pm 0.316$ & $0.598 \pm 0.022$ \\
    Fine-tuning & All 6 & $663{,}552$ & $\underline{0.382 \pm 0.012}$ & $0.807 \pm 0.020$ & $1.576 \pm 0.028$ & $0.670 \pm 0.012$ \\
    \midrule
    PGS (ours) & Bottom-up 1 & $\mathbf{192}$ & $\mathbf{0.417 \pm 0.007}$ & $0.765 \pm 0.014$ & $1.526 \pm 0.026$ & $0.719 \pm 0.009$ \\
    PGS (ours) & Bottom-up 2 & $\underline{384}$ & $0.399 \pm 0.006$ & $0.759 \pm 0.012$ & $\underline{1.545 \pm 0.025}$ & $0.683 \pm 0.011$ \\
    PGS (ours) & Bottom-up 4 & $768$ & $\underline{0.405 \pm 0.006}$ & $\underline{0.796 \pm 0.015}$ & $1.542 \pm 0.013$ & $0.686 \pm 0.016$ \\
    PGS (ours) & Top-down 1 & $\mathbf{192}$ & $0.152 \pm 0.010$ & $0.104 \pm 0.039$ & $0.448 \pm 0.182$ & $0.638 \pm 0.033$ \\
    PGS (ours) & Top-down 2 & $\underline{384}$ & $0.194 \pm 0.008$ & $0.300 \pm 0.023$ & $1.133 \pm 0.104$ & $\mathbf{0.626 \pm 0.028}$ \\
    PGS (ours) & Top-down 4 & $768$ & $0.354 \pm 0.011$ & $0.725 \pm 0.019$ & $\mathbf{1.546 \pm 0.037}$ & $\underline{0.632 \pm 0.026}$ \\
    PGS (ours) & Head & $\mathbf{192}$ & $0.093 \pm 0.007$ & $-0.082 \pm 0.019$ & $-1.477 \pm 0.369$ & $0.640 \pm 0.014$ \\
    PGS (ours) & All 6 & $1{,}152$ & $0.403 \pm 0.005$ & $\mathbf{0.805 \pm 0.014}$ & $1.527 \pm 0.011$ & $0.690 \pm 0.019$ \\
    \bottomrule
  \end{tabular}%
  }
  \caption{Layer-support ablation across the four matched methods. Bottom-up support starts at the earliest residual block, top-down support starts at the latest block, and every independently fitted object receives one total isolated action-KL budget of $0.1$.}
  \label{rethink:tab:chess-layers}
\end{table}

\paragraph{Fit-data budget.}
The nested data-budget study shows a clear low-data regime with weaker and
more variable composition.  Isolated gain, composed gain, and retention
stabilize as the fit set grows, with no material composed-gain improvement
from doubling the promoted budget.  This supports 40 puzzles per motif as a
reasonable fixed default without selecting it on test outcomes.

\begin{table}[H]
  \centering
  \scriptsize
  \setlength{\tabcolsep}{2.2pt}
  \adjustbox{max width=\linewidth}{%
  \begin{tabular}{lrrrr}
    \toprule
    Puzzles/motif & Isolated gain $\uparrow$ & Composed gain $\uparrow$ & Min.\ retention $\uparrow$ & Composed KL $\downarrow$ \\
    \midrule
    5 & $0.360 \pm 0.030$ & $0.725 \pm 0.082$ & $1.353 \pm 0.249$ & $\mathbf{0.583 \pm 0.061}$ \\
    10 & $0.385 \pm 0.011$ & $0.785 \pm 0.017$ & $1.453 \pm 0.081$ & $\underline{0.642 \pm 0.035}$ \\
    20 & $0.396 \pm 0.007$ & $0.793 \pm 0.013$ & $1.478 \pm 0.056$ & $0.674 \pm 0.028$ \\
    40 & $\underline{0.403 \pm 0.005}$ & $\mathbf{0.805 \pm 0.014}$ & $\underline{1.527 \pm 0.011}$ & $0.690 \pm 0.019$ \\
    80 & $\mathbf{0.404 \pm 0.006}$ & $\underline{0.801 \pm 0.014}$ & $\mathbf{1.532 \pm 0.018}$ & $0.702 \pm 0.006$ \\
    \bottomrule
  \end{tabular}%
  }
  \caption{PGS fit-data ablation using nested puzzle subsets (each smaller budget is contained in the next) with the promoted all-six-layer channel parameterization. The 40-puzzle setting is the frozen default; test outcomes do not select the budget.}
  \label{rethink:tab:chess-data-budget}
\end{table}

\paragraph{Optimization steps.}
Additional PGS updates trade isolated performance and policy-change efficiency
for stronger three-way composition.  ReFT and LoRA improve more gradually,
whereas Fine-tuning is effectively unchanged across the tested update counts.
The promoted one-update PGS configuration therefore represents the balanced
isolated/composed setting rather than the maximum-composition setting.  The
baseline defaults remain frozen at three updates.

\begin{table}[H]
  \centering
  \scriptsize
  \setlength{\tabcolsep}{2.2pt}
  \adjustbox{max width=\linewidth}{%
  \begin{tabular}{llrrrrr}
    \toprule
    Updates & Method & Scalars $\downarrow$ & Isolated gain $\uparrow$ & Composed gain $\uparrow$ & Min.\ retention $\uparrow$ & Composed KL $\downarrow$ \\
    \midrule
    1 & ReFT & $\underline{9{,}288}$ & $0.371 \pm 0.011$ & $0.740 \pm 0.013$ & $1.475 \pm 0.047$ & $\mathbf{0.602 \pm 0.024}$ \\
    1 & LoRA & $46{,}080$ & $0.362 \pm 0.015$ & $0.761 \pm 0.038$ & $\underline{1.544 \pm 0.079}$ & $\underline{0.613 \pm 0.032}$ \\
    1 & Fine-tuning & $663{,}552$ & $\underline{0.382 \pm 0.011}$ & $\mathbf{0.807 \pm 0.020}$ & $\mathbf{1.576 \pm 0.027}$ & $0.670 \pm 0.012$ \\
    \midrule
    1 & PGS (ours) & $\mathbf{1{,}152}$ & $\mathbf{0.403 \pm 0.005}$ & $\underline{0.805 \pm 0.014}$ & $1.527 \pm 0.011$ & $0.690 \pm 0.019$ \\
    \midrule
    3 & ReFT & $\underline{9{,}288}$ & $0.379 \pm 0.004$ & $0.759 \pm 0.006$ & $1.489 \pm 0.023$ & $\underline{0.667 \pm 0.012}$ \\
    3 & LoRA & $46{,}080$ & $0.377 \pm 0.010$ & $0.799 \pm 0.023$ & $\mathbf{1.581 \pm 0.056}$ & $\mathbf{0.654 \pm 0.014}$ \\
    3 & Fine-tuning & $663{,}552$ & $\underline{0.382 \pm 0.012}$ & $\underline{0.807 \pm 0.020}$ & $\underline{1.576 \pm 0.028}$ & $0.670 \pm 0.012$ \\
    \midrule
    3 & PGS (ours) & $\mathbf{1{,}152}$ & $\mathbf{0.400 \pm 0.004}$ & $\mathbf{0.827 \pm 0.007}$ & $1.561 \pm 0.025$ & $0.753 \pm 0.013$ \\
    \midrule
    10 & ReFT & $\underline{9{,}288}$ & $\mathbf{0.385 \pm 0.001}$ & $0.780 \pm 0.007$ & $1.504 \pm 0.022$ & $0.710 \pm 0.019$ \\
    10 & LoRA & $46{,}080$ & $0.378 \pm 0.012$ & $0.807 \pm 0.026$ & $\underline{1.588 \pm 0.044}$ & $\underline{0.673 \pm 0.012}$ \\
    10 & Fine-tuning & $663{,}552$ & $\underline{0.382 \pm 0.012}$ & $\underline{0.807 \pm 0.021}$ & $1.575 \pm 0.028$ & $\mathbf{0.671 \pm 0.012}$ \\
    \midrule
    10 & PGS (ours) & $\mathbf{1{,}152}$ & $0.380 \pm 0.012$ & $\mathbf{0.852 \pm 0.017}$ & $\mathbf{1.599 \pm 0.019}$ & $0.785 \pm 0.020$ \\
    \bottomrule
  \end{tabular}%
  }
  \caption{Update-count sensitivity on matched all-six-layer support. The promoted defaults are one update for PGS and three updates for ReFT, LoRA, and Fine-tuning.}
  \label{rethink:tab:chess-updates}
\end{table}

\paragraph{PGS parameterization.}
Channel PGS is both substantially smaller and behaviorally stronger than the
spatial control in isolated and composed gain.  Spatial PGS induces a smaller
three-way KL, but this accompanies much weaker target improvement rather than
better composition at matched effectiveness.  We therefore retain channel
PGS as the canonical parameterization.

\begin{table}[H]
  \centering
  \scriptsize
  \setlength{\tabcolsep}{2.2pt}
  \adjustbox{max width=\linewidth}{%
  \begin{tabular}{lrrrrr}
    \toprule
    PGS parameterization & Scalars $\downarrow$ & Isolated gain $\uparrow$ & Composed gain $\uparrow$ & Min.\ retention $\uparrow$ & Composed KL $\downarrow$ \\
    \midrule
    Channel & $\mathbf{1{,}152}$ & $\mathbf{0.403 \pm 0.005}$ & $\mathbf{0.805 \pm 0.014}$ & $\mathbf{1.527 \pm 0.011}$ & $\underline{0.690 \pm 0.019}$ \\
    Spatial & $\underline{73{,}728}$ & $\underline{0.208 \pm 0.011}$ & $\underline{0.443 \pm 0.035}$ & $\underline{1.313 \pm 0.206}$ & $\mathbf{0.440 \pm 0.045}$ \\
    \bottomrule
  \end{tabular}%
  }
  \caption{PGS parameterization control over all six residual blocks. Channel PGS stores one broadcast coefficient per channel and site; spatial PGS stores a coefficient at every channel and board location.}
  \label{rethink:tab:chess-parameterization}
\end{table}

\section{Additional Football Details}
\label{rethink:app:football}

\subsection{Protocol}

\begin{itemize}
  \item Collect complete unsteered matches against a source opponent and fit
  PGS from the controlled policy's actions and returns.
  \item Keep fit, validation, and evaluation seeds disjoint; validation fixes
  the operating point under an action-KL budget.
  \item Apply the same frozen artifact against its source and unseen opponents.
  \item Report paired held-out goal difference, action KL, and event-based
  behavior measures, with base and KL-matched random controls.
\end{itemize}

\paragraph{Behavior Objectives.}
The main football study fixes three outcomes: completed passes, possession
regains, and shot attempts. The implementation also supports possession,
turnover avoidance, and wide attacking play as configurable objectives. These
alternatives are protocol options rather than main-study outcomes; changing the
objective requires a separately specified fit, validation, and held-out
evaluation.

\subsection{Player-Role Structure and Base Phenotype}

\begin{figure}[H]
  \centering
  \includegraphics[width=0.82\linewidth]{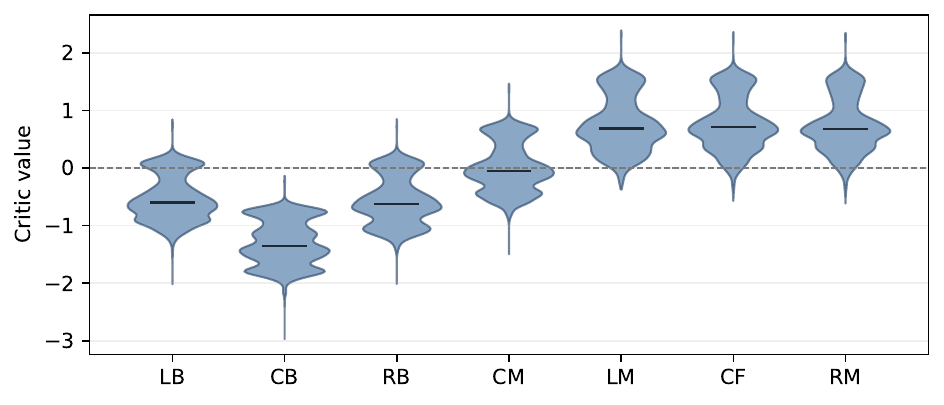}
  \caption{Frozen critic values for \texttt{flank\_pass}, grouped by the active
  player's role. The distributions reveal a pronounced positional bias in the
  uncentered critic output.}
  \label{rethink:fig:football-critic-role-bias}
\end{figure}

\paragraph{Critic Player-Role Bias.}
The frozen critic's values vary systematically with the active player's role
in \texttt{flank\_pass} (Figure~\ref{rethink:fig:football-critic-role-bias}).
Median values rise from center backs through fullbacks and central midfielders
to wide midfielders and center forwards. Raw critic magnitude therefore
contains a strong role-dependent offset and is not, by itself, evidence of
opponent-specific structure. We remove this offset by centering critic values
within each player role before constructing critic-weighted directions.

\paragraph{Behavior Definitions and Base Phenotype.}
Table~\ref{tab:football-appendix-behavior-description} reports the four
additional unsteered full-match behaviors over five self-play matches per
policy: possession rate, turnovers, final-third entries, and wide final-third
entries. We compute
these as possible behavioral targets.

\begin{table}[H]
  \centering
  \scriptsize
  \adjustbox{max width=\linewidth}{%
  \begin{tabular}{lrrrr}
    \toprule
    Self-play policy & Possession share $\uparrow$ & Turnovers $\downarrow$ & Final-third entries $\uparrow$ & Wide final-third entries $\uparrow$ \\
    \midrule
    \texttt{defensive\_passer} & 0.21 $\pm$ 0.01 & \textbf{26.20 $\pm$ 2.59} & \textbf{5.40 $\pm$ 0.89} & \textbf{4.60 $\pm$ 0.55} \\
    \texttt{flank\_pass} & 0.21 $\pm$ 0.03 & \underline{36.40 $\pm$ 4.34} & \underline{5.00 $\pm$ 1.87} & \underline{3.00 $\pm$ 1.22} \\
    \texttt{group\_pressure} & \textbf{0.34 $\pm$ 0.04} & 91.40 $\pm$ 13.37 & 5.00 $\pm$ 1.22 & 2.40 $\pm$ 1.14 \\
    \texttt{offensive\_passer} & \underline{0.22 $\pm$ 0.03} & 41.20 $\pm$ 4.49 & 5.00 $\pm$ 2.92 & 0.60 $\pm$ 0.89 \\
    \bottomrule
  \end{tabular}%
  }
  \caption{Unsteered full-match appendix behaviors; values are mean $\pm$ sample standard deviation over five self-play matches.}
  \label{tab:football-appendix-behavior-description}
\end{table}

\subsection{Opponent-Conditioned Adaptation}
\label{rethink:sec:football-adversary}

This study asks whether prior interaction with an opponent can produce a useful
temporary response from the frozen \texttt{group\_pressure} policy. PGS fits one
temporary intervention from \texttt{group\_pressure} trajectories collected
against \texttt{defensive\_passer}, then evaluates that intervention both
against \texttt{defensive\_passer} and in self-play. Thus, opponent-conditioned
describes the interaction data that shape the intervention, rather than an
explicit learned model of the opponent.

\begin{table}[H]
  \centering
  \small
  \setlength{\tabcolsep}{4.5pt}
  \adjustbox{max width=\columnwidth}{%
  \begin{tabular}{llrr}
    \toprule
    Evaluation & Method & Goal diff. $\uparrow$ & KL $\downarrow$ \\
    \midrule
    defensive\_passer & Base & \ensuremath{-1.00 \pm 1.22} & -- \\
    defensive\_passer & PGS (ours) & \ensuremath{+0.20 \pm 0.84} & \ensuremath{+0.102 \pm 0.010} \\
    \midrule
    Self-play & Base & \ensuremath{+0.20 \pm 0.45} & -- \\
    Self-play & PGS (ours) & \ensuremath{-0.40 \pm 0.55} & \ensuremath{+0.126 \pm 0.007} \\
    \bottomrule
  \end{tabular}%
  }
  \caption{Opponent-conditioned goal steering. Values are mean $\pm$ sample standard deviation over held-out match seeds.}
  \label{tab:football-opponent-conditioned-goals}
\end{table}

\subsection{Ablations}

\paragraph{Steering from Off-Policy Trajectories.}
Table~\ref{tab:football-appendix-off-policy} steers the frozen
\texttt{group\_pressure} policy using trajectories demonstrated by
\texttt{defensive\_passer} against \texttt{group\_pressure}. It compares a
ratio-free off-policy PGS surrogate with the same surrogate weighted by the
per-action likelihood ratio. This policy-pair-specific diagnostic tests
sensitivity to ratio weighting, rather than establishing an unbiased off-policy
estimator or a general football advantage.

\begin{table}[H]
  \centering
  \small
  \setlength{\tabcolsep}{4.5pt}
  \adjustbox{max width=\columnwidth}{%
  \begin{tabular}{lrrr}
    \toprule
    Method & Completed passes & Goal diff. & KL \\
    \midrule
    Base & \ensuremath{+9.00 \pm 4.90} & \ensuremath{-0.60 \pm 0.55} & -- \\
    PGS (no ratio) & \ensuremath{\underline{+37.20 \pm 13.24}} & \ensuremath{-0.80 \pm 1.30} & \ensuremath{\underline{+0.105 \pm 0.008}} \\
    PGS (ours) & \ensuremath{\mathbf{+48.20 \pm 16.89}} & \ensuremath{+0.00 \pm 0.71} & \ensuremath{\mathbf{+0.096 \pm 0.011}} \\
    \bottomrule
  \end{tabular}%
  }
  \caption{Off-policy completed-passes steering. Values are mean $\pm$ sample standard deviation.}
  \label{tab:football-appendix-off-policy}
\end{table}

\paragraph{Critic-Based Credit Assignment.}
We compare five simple baselines for completed-pass returns: no centering,
centering over the fit trajectory, centering by player type within each
collected batch, centering by player type after pooling the fit batches, and
subtraction of an observation-based ridge predictor fitted on the fit
trajectories. The last condition is an in-sample learned baseline, not a
cross-fitted or causal critic.

\begin{table}[H]
  \centering
  \small
  \setlength{\tabcolsep}{4.5pt}
  \adjustbox{max width=\columnwidth}{%
  \begin{tabular}{lrrr}
    \toprule
    Fit-weight baseline & Completed passes $\uparrow$ & Goal diff. $\uparrow$ & Action KL $\downarrow$ \\
    \midrule
    Base & \ensuremath{+1.80 \pm 3.49} & \ensuremath{+0.00 \pm 0.00} & -- \\
    None & \ensuremath{+15.20 \pm 14.20} & \ensuremath{\mathbf{+0.00 \pm 0.00}} & \ensuremath{\underline{+0.103 \pm 0.014}} \\
    Trajectory centered & \ensuremath{\underline{+25.60 \pm 16.16}} & \ensuremath{\mathbf{+0.00 \pm 0.71}} & \ensuremath{+0.106 \pm 0.018} \\
    Player-type centered & \ensuremath{\underline{+25.60 \pm 18.66}} & \ensuremath{\mathbf{+0.00 \pm 0.00}} & \ensuremath{+0.106 \pm 0.013} \\
    Player-type centered (pooled) & \ensuremath{\mathbf{+27.80 \pm 20.02}} & \ensuremath{\underline{-0.20 \pm 0.45}} & \ensuremath{+0.107 \pm 0.015} \\
    Learned critic from obs. & \ensuremath{+18.20 \pm 18.89} & \ensuremath{\mathbf{+0.00 \pm 0.00}} & \ensuremath{\mathbf{+0.102 \pm 0.019}} \\
    \bottomrule
  \end{tabular}%
  }
  \caption{Completed-passes steering under simple reward-weight baselines. Values are mean $\pm$ sample standard deviation over held-out match seeds.}
  \label{tab:football-appendix-critic}
\end{table}

\paragraph{Role-Specific Steering.}
We compare a shared intervention with independently routed vectors by player
role and player index. Because these schemes allocate different numbers of
vectors, this is a routing-and-capacity ablation rather than a pure test of role
semantics.

\begin{table}[H]
  \centering
  \small
  \setlength{\tabcolsep}{4.5pt}
  \adjustbox{max width=\columnwidth}{%
  \begin{tabular}{lrrr}
    \toprule
    Method & Completed passes $\uparrow$ & Goal diff. $\uparrow$ & KL $\downarrow$ \\
    \midrule
    Base & \ensuremath{+2.20 \pm 3.03} & \ensuremath{-0.40 \pm 0.89} & -- \\
    shared PGS (ours) & \ensuremath{\underline{+17.20 \pm 12.83}} & \ensuremath{\underline{-0.60 \pm 0.89}} & \ensuremath{\mathbf{+0.107 \pm 0.017}} \\
    role PGS (ours) & \ensuremath{\mathbf{+26.00 \pm 10.91}} & \ensuremath{\mathbf{-0.40 \pm 0.89}} & \ensuremath{\underline{+0.114 \pm 0.003}} \\
    agent PGS (ours) & \ensuremath{+14.80 \pm 7.40} & \ensuremath{\underline{-0.60 \pm 0.55}} & \ensuremath{+0.118 \pm 0.006} \\
    \bottomrule
  \end{tabular}%
  }
  \caption{Held-out completed passes steering. Values are mean $\pm$ sample standard deviation over held-out match seeds.}
  \label{tab:football-appendix-role}
\end{table}

\paragraph{Fit-Data Budget.}
Table~\ref{tab:football-appendix-data-budget} varies the self-play data used
to fit a shared completed-passes intervention for \texttt{group\_pressure}.
The split-data condition retains equal temporal halves from the left and right
policy trajectories. This descriptive ablation does not establish monotonic
scaling or a confirmed behavioral effect.

\begin{table}[H]
  \centering
  \small
  \setlength{\tabcolsep}{4.5pt}
  \adjustbox{max width=\columnwidth}{%
  \begin{tabular}{llrrr}
    \toprule
    Method & fit\_match\_budget & Completed passes & Goal diff. & KL \\
    \midrule
    Base &  & \ensuremath{+5.20 \pm 4.32} & \ensuremath{-0.80 \pm 0.84} & -- \\
    PGS (ours) & 0.5 & \ensuremath{+16.00 \pm 9.27} & \ensuremath{-0.20 \pm 0.84} & \ensuremath{\underline{+0.102 \pm 0.011}} \\
    PGS (ours) & 1 & \ensuremath{+14.60 \pm 9.07} & \ensuremath{-0.20 \pm 0.45} & \ensuremath{\mathbf{+0.099 \pm 0.004}} \\
    PGS (ours) & 3 & \ensuremath{\mathbf{+32.60 \pm 24.50}} & \ensuremath{-0.40 \pm 0.89} & \ensuremath{+0.113 \pm 0.017} \\
    PGS (ours) & 5 & \ensuremath{\underline{+28.40 \pm 22.22}} & \ensuremath{+0.00 \pm 0.71} & \ensuremath{+0.118 \pm 0.012} \\
    \bottomrule
  \end{tabular}%
  }
  \caption{Fit-data-budget ablation. Values are mean $\pm$ sample standard deviation.}
  \label{tab:football-appendix-data-budget}
\end{table}

\paragraph{Discount Factor.}
Table~\ref{tab:football-appendix-gamma} varies the return discount factor while
holding the self-play fit budget at one match (both policy sides; 6,000 steps).
This descriptive sweep does not select a discount factor or establish a
confirmed behavioral effect.

\begin{table}[H]
  \centering
  \small
  \setlength{\tabcolsep}{4.5pt}
  \adjustbox{max width=\columnwidth}{%
  \begin{tabular}{llrrr}
    \toprule
    Method & gamma & Completed passes & Goal diff. & KL \\
    \midrule
    Base &  & \ensuremath{+7.40 \pm 7.44} & \ensuremath{+0.20 \pm 0.45} & -- \\
    PGS (ours) & 0.0 & \ensuremath{+8.80 \pm 9.01} & \ensuremath{+0.20 \pm 0.45} & \ensuremath{+0.106 \pm 0.009} \\
    PGS (ours) & 0.25 & \ensuremath{\mathbf{+29.80 \pm 15.37}} & \ensuremath{+0.00 \pm 0.71} & \ensuremath{+0.111 \pm 0.007} \\
    PGS (ours) & 0.5 & \ensuremath{+18.20 \pm 11.67} & \ensuremath{+0.00 \pm 1.22} & \ensuremath{+0.116 \pm 0.017} \\
    PGS (ours) & 0.8 & \ensuremath{\underline{+19.60 \pm 11.50}} & \ensuremath{+0.00 \pm 0.00} & \ensuremath{\underline{+0.105 \pm 0.012}} \\
    PGS (ours) & 0.9 & \ensuremath{+8.40 \pm 7.40} & \ensuremath{+0.00 \pm 0.00} & \ensuremath{+0.106 \pm 0.001} \\
    PGS (ours) & 0.99 & \ensuremath{+9.00 \pm 8.25} & \ensuremath{-0.60 \pm 0.89} & \ensuremath{+0.116 \pm 0.020} \\
    PGS (ours) & 1.0 & \ensuremath{+7.20 \pm 3.77} & \ensuremath{+0.40 \pm 0.89} & \ensuremath{\mathbf{+0.103 \pm 0.016}} \\
    \bottomrule
  \end{tabular}%
  }
  \caption{Discount-factor ablation. Values are mean $\pm$ sample standard deviation.}
  \label{tab:football-appendix-gamma}
\end{table}

\section{Activation-Site Calibration and Sensitivity}
\label{rethink:app:activation-sensitivity}

\paragraph{Euclidean sensitivity.}
Let $J_l(h_l)$ denote the PGS objective downstream of a candidate activation
site and let $q_l=\nabla_{h_l}J_l$.  For an additive intervention
$\delta_l=\alpha q_l$, a first-order expansion predicts
\begin{equation}
  J_l(h_l+\delta_l)-J_l(h_l)
  \approx \alpha\norm{q_l}^2.
  \label{rethink:eq:activation-loss-sensitivity}
\end{equation}
The gradient norm is therefore a cheap loss-sensitivity screen, but it is not
a complete layer-selection rule: activation scale, dimensionality, curvature,
and downstream policy effects differ across sites.

\paragraph{Fixed-direction calibration.}
Let $F_l$ be the activation-space empirical Fisher and write
$\pi_{\alpha,l}$ for the policy induced by $\delta_l=\alpha q_l$. A local
quadratic approximation to its action KL gives
\begin{equation}
  \begin{aligned}
    D_{\mathrm{KL}}(\pi,\pi_{\alpha,l})
    &\approx \tfrac12\alpha^2 q_l^\top F_lq_l,\\
    \alpha_l
    &=\sqrt{\frac{2\varepsilon}{q_l^\top F_lq_l}},
  \end{aligned}
  \label{rethink:eq:activation-fixed-direction-scale}
\end{equation}
where the second line sets the approximation equal to the policy-change
budget $\varepsilon$. This is the fixed-direction calibration used by PGS.

\paragraph{Policy-aware site sensitivity.}
A more general policy-aware site diagnostic also allows the intervention
direction to change. The KL-constrained local problem and its solution are
\begin{equation}
  \begin{aligned}
    &\max_{\delta_l}\ q_l^\top\delta_l
    &&\text{s.t.}\quad
      \tfrac12\delta_l^\top F_l\delta_l\leq\varepsilon,\\
    &\delta_l^*
    &&=\sqrt{\frac{2\varepsilon}{q_l^\top F_l^{-1}q_l}}
      F_l^{-1}q_l.
  \end{aligned}
  \label{rethink:eq:activation-fisher-sensitivity}
\end{equation}
Thus $q_l^\top F_l^{-1}q_l$ is the predicted objective gain per local
policy-change budget. Both site-sensitivity criteria are diagnostics only: the
intervention site and finite strength are selected on validation data, and the
final action KL and behavioral effect are measured on held-out data.

\section{Reproducibility}
\label{rethink:app:reproducibility}

\subsection{Code and Data}

Reusable implementations, experiment entrypoints, and frozen
configurations are under \texttt{src/}, \texttt{scripts/}, and
\texttt{configs/}.  The configurations specify the model or environment,
data split, intervention and calibration settings, and all random, fit, and
partition seeds.  For example, the chess experiments use fit seeds 42--46
with a fixed partition seed (20260723); comparisons share the corresponding
splits and evaluation seeds.  Entry points write per-seed diagnostics and
machine-readable result rows, from which the paper tables are generated.

\subsection{Software}

Experiments use Python 3.11 with dependencies managed by
\texttt{uv}~\citep{uv}.  The numerical, configuration, and plotting stack uses
NumPy, Hydra, and Matplotlib~\citep{Harris2020ArrayPW,Yadan2019Hydra,
Hunter2007MatplotlibA2}.  Neural-policy experiments use
PyTorch~\citep{Ansel2024PyTorch2F}; chess experiments use
\texttt{lczerolens}~\citep{poupart_lczerolens_2026}; football experiments use
Google Research Football~\citep{Kurach2019GoogleRF}.

\subsection{Hardware}

The GPU launches use one NVIDIA Tesla V100 GPU, 10 CPU cores, and 40~GB
of host memory per GPU job.  Lightweight diagnostic workflows are run
locally.

\end{document}